\documentclass[11pt]{article}

\usepackage[table]{xcolor}

\usepackage[final]{acl}

\usepackage{times}
\usepackage{latexsym}
\usepackage{hyperref}
\usepackage{url}
\usepackage{graphicx}

\usepackage{amsmath}
\usepackage{amssymb}
\usepackage{booktabs}
\usepackage{makecell}
\usepackage{multirow}
\usepackage{placeins}
\usepackage{pdflscape}
\usepackage{tikz}
\usetikzlibrary{positioning, arrows.meta, calc, fit, shapes.geometric}
\usepackage[normalem]{ulem}  
\usepackage[T1]{fontenc}

\usepackage[utf8]{inputenc}

\usepackage{CJKutf8}
\newcommand{\zh}[1]{\begin{CJK*}{UTF8}{gbsn}#1\end{CJK*}}
\newcommand{\ja}[1]{\begin{CJK}{UTF8}{min}#1\end{CJK}} 

\usepackage{microtype}

\usepackage{inconsolata}

\title{Cross-Lingual Alignment Without Joint Training: \\
Do Monolingual Language Models Converge on Universal Representations?}

\author{
Ej Zhou$^{1}$ \quad Suchir Salhan$^{1}$ \quad Catherine Arnett$^{2}$ \quad Anna Korhonen$^{1}$ \\
$^{1}$University of Cambridge \quad $^{2}$EleutherAI \\
\texttt{\{yz926,sas245,alk23\}@cam.ac.uk}  \quad \texttt{catherine@eleuther.com}
}

\begin{document}

\maketitle

\begin{abstract}

Cross-lingual alignment in multilingual language models is what enables cross-lingual transfer and is typically attributed to joint training: shared parameters, mixed-language batches, or explicit alignment objectives. 
In this paper, we ask whether monolingual models trained on non-parallel data learn alignable representations without joint training.
By testing on strictly monolingual language models, such as the Goldfish model families and independently developed models from different research labs, we find three results. \emph{Correlation}: these models develop alignable representational geometry across layers, with alignment strengthening as data scale, model scale, or linguistic proximity increases. \emph{Construction}: a single Procrustes rotation fit on parallel sentences maps hidden states between models. \emph{Causation}: the same rotation transfers functional content; patching a rotated English residual into a German model on a factual cloze task flips the prediction to the English model's answer in most cases. We argue that cross-lingual alignment can emerge from the structure of language and the information it carries rather than from joint training and shared representations. This points to practical future directions including model stitching, merging, and modular multilingual systems built from monolingual components.

\end{abstract}

\section{Introduction}

Multilingual language models produce representations in which cross-lingual semantically parallel inputs occupy close regions of activation space \citep{chang-etal-2022-geometry, brinkmann-etal-2025-large}. The research community has proposed various explanations in mechanism: shared subword vocabularies \citep{pires2019multilingual}, similar input distributions \citep{conneau2020emerging}, or shared architectural parameters \citep{dufter-schutze-2020-identifying}. These studies only investigate the cause within the features of joint training, and no existing account predicts cross-lingual alignment between models trained independently on disjoint corpora.

In this paper, we ask whether joint training is required for cross-lingual alignment. We pose this question in its strictest form: if two models are trained independently, on distinct corpora and in different languages, with no shared parameters or alignment signal, do their representations still end up alignable? A positive answer would attest the emergence of language-agnostic conceptual representations purely from the corpora. This is predicted by the Platonic Representation Hypothesis \citep{huh2024position}, which has been tested cross-modally but not cross-linguistically. In our experiments, we consider only strictly monolingual language models: each is a Transformer trained on a single language. Our central findings replicate on five independently developed ${\sim}$1B-parameter monolingual models from different research groups, showing the effect holds at larger scale.

\begin{figure*}[t]
\centering
\includegraphics[width=\textwidth, trim=0 110 0 0,clip]{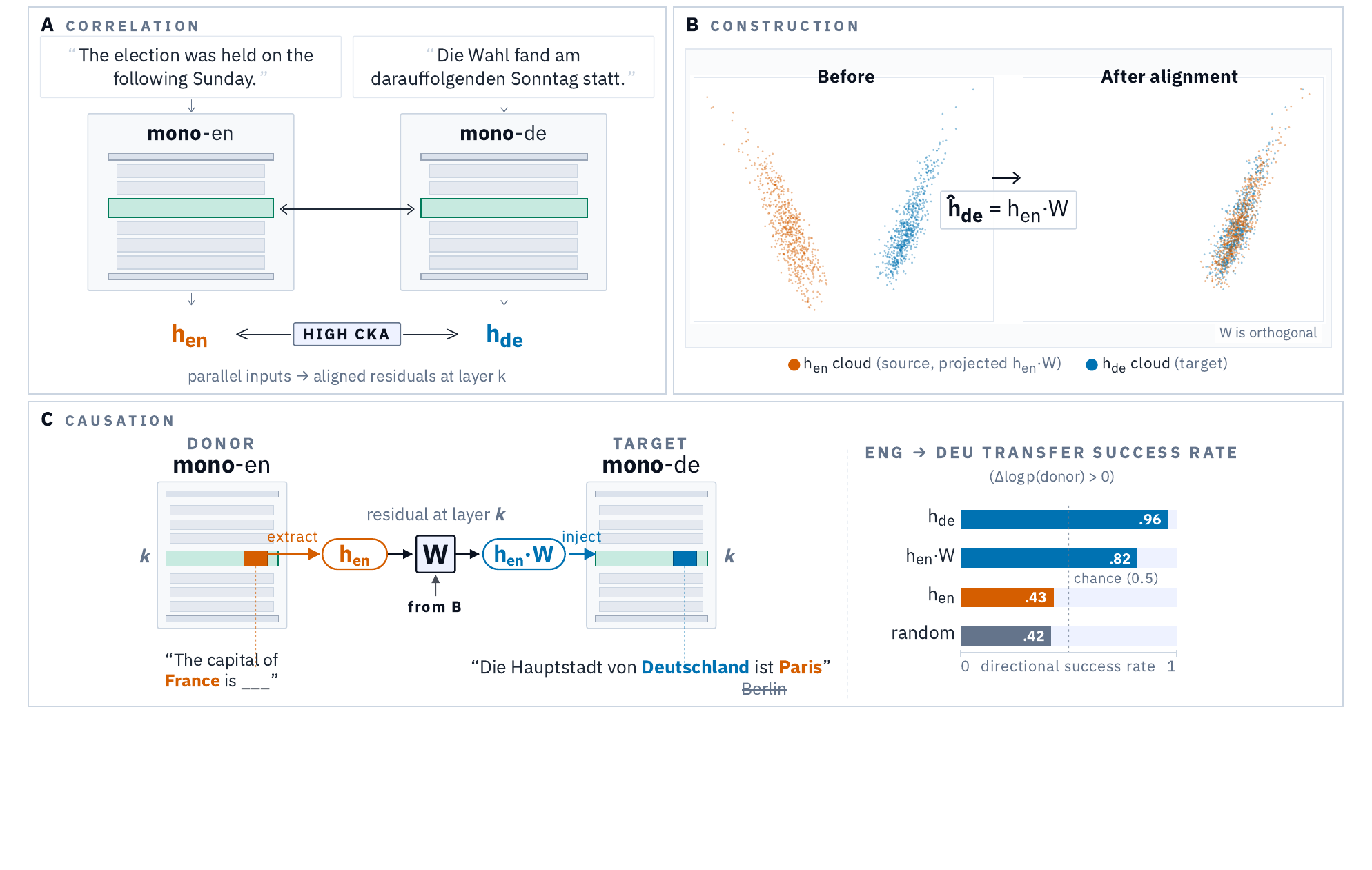}
\caption{\textbf{Independent monolingual LMs share representational
geometry; a rotation makes that geometry causally usable.}
\textbf{(A)}~Parallel English/German sentences fed to monolingual models yield residual streams $h_{\text{en}}$,
$h_{\text{de}}$ with high CKA.
\textbf{(B)}~A Procrustes rotation $W$ fit on parallel sentence
representations aligns the two clouds: $\widehat{Y}{=}XW$ overlaps the target.
\textbf{(C)}~The same $W$ transfers factual content across models.
Patching $W{\cdot}h_{\text{en}}$ at the residual of a
\textsc{mono-de} prompt flips the German completion, recovering directional success rate ($\Delta\log p(\text{donor}){>}0$)
 most of the ceiling and well above baselines.  
 }
\label{fig:teaser}
\end{figure*}

Our analysis is three-part, with increasingly stringent standards of evidence: \emph{\textbf{C}orrelation, \textbf{C}onstruction, and \textbf{C}ausation}. In the correlational experiments (Section~\ref{sec:representational_alignment}) we find that independently trained monolingual models develop alignable representational geometry on parallel sentences, with alignment strengthening as data scale, model scale, or linguistic proximity increases, and persisting across models from different research groups. In the constructional experiments (Section~\ref{sec:reconstruction}), we compare an orthogonal rotation against more flexible mappings (affine, MLP) and find that a single rotation is sufficient to align one model's representations with another's. The orthogonality constraint preserves the angular geometry that retrieval depends on, whereas the relaxed alternatives lower reconstruction error but weaken retrieval. Finally, in the causal experiments (Section~\ref{sec:crossmodel_patching}), we show that the same rotation transfers functional content: patching a rotated English residual into a German model on a factual cloze task flips the prediction to the donor's capital in most cases, with the pattern holding across five unrelated target languages and approaching the within-model ceiling.\footnote{Code and data for all experiments: \url{https://github.com/Ej-zhou/monolingual-alignment}}

\section{Related Work}

\subsection{Cross-Lingual Representations in Multilingual LMs}

A growing literature finds that multilingual language models develop partly shared representational structure across languages. Early evidence in jointly trained encoders established that languages occupy overlapping linear subspaces \citep{chang-etal-2022-geometry}. Subsequent work on decoder-only multilingual LMs has found a similarly partial language-agnostic geometry, with internal components shared across languages while still allowing language-specific divergences \citep{ICLR2025_edd00cea, wendler2024llamas, brinkmann-etal-2025-large,
dumas-etal-2025-separating, riemenschneider-frank-2025-cross,
korner-etal-2026-meanings, harrasse2026tracingmultilingualrepresentationsllms}. More recently, the Platonic Representation Hypothesis \citep{huh2024position} posits that representations converge towards a shared model of reality as models scale. \citet{jha2026harnessing} support this at the embedding level, showing text embeddings can be translated across architectures without paired data. In this paper, we connect these predictions and ask whether there are cross-lingually universal representations learned by monolingual models. 

\subsection{Monolingual LMs}
Most state-of-the-art large-scale LMs \citep{qwen2025qwen25technicalreport,gemmateam2024gemma2improvingopen} and well-studied multilingual LMs \citep{devlin2019bert,conneau2020xlmr,lescao2023bloom} are trained jointly across many languages. For our experiments, we require strictly monolingual models, in order to study whether language-agnostic structure can emerge independently of multilingual joint training. Recent efforts have produced such models, namely the Goldfish models \citep{chang2024goldfish},  which include comparably trained monolingual models for 350 languages. These provide a controlled setting for our experiments.  There are also several monolingual models developed by language communities (e.g. Minerva, \citealp{orlando-etal-2024-minerva}; Zh-Pythia \citealp[]{liu2025systematicassessmentlanguagemodels}; Tucano, \citealp[]{correa2025tucanoadvancingneuraltext}; and Bielik, \citealp[]{ociepa2025bielikv3smalltechnical}).
We use these models to test the scalability of our findings, albeit in a less controlled setting.

\subsection{Aligning Independently-Trained LMs}
\label{sec:related-monolingual}

Several studies have found that independently trained representations can be aligned \textit{post hoc}. Early work in Bilingual Lexicon Induction (BLI) demonstrates that monolingual word embedding spaces are often related through approximately orthogonal transformations, enabling bilingual lexicon induction without parallel supervision \citep{mikolov2013exploiting, xing-etal-2015-normalized, smith2017offline, artetxe-etal-2018-robust, conneau2018word}. Orthogonality constraints usually improve stability and performance relative to unconstrained linear mappings, suggesting that independently trained semantic spaces preserve similar global geometry. Geometric alignment measures, such as Centered Kernel Alignment \citep{kornblith2019similarity}, which builds on the Hilbert--Schmidt Independence Criterion (HSIC; \citealp{gretton2005measuring}), a kernel-based measure of statistical dependence between two sets of representations, have been used to probe multilingual encoders \citep{muller2021first}. However, correlational similarity is insufficient on its own for making claims about shared representation, as CKA can be dominated by a small number of high-variance components \citep{davari2023reliability}, inflated by model width and depth with only local neighbourhood agreement surviving null calibration \citep{groger2026revisiting}. Additionally,  nearest-neighbour alignment (as was used in \citealp{huh2024position}) can be fragile to evaluation scale \citep{koepke2026back}. Mechanistic interpretability approaches provide a stronger test of representational convergence by testing functional interchangeability: whether aligned states are usable by the receiving model's computation. In this work, we use cross-model activation patching \citep{dumas-etal-2025-separating,korner-etal-2026-meanings} as a causal test. If a representation from one model can be projected into another model's residual stream and successfully drive downstream behaviour on a new task, then the alignment must preserve functionally relevant structure rather than correlational similarity alone.

The closest prior work to ours is \citet{conneau2020emerging}, who train monolingual BERT models on separate corpora, measure their similarity with CKA, and align them \textit{post hoc} with a learned linear map. We extend their findings in four directions: (i) we analyse decoder-only causal LMs rather than masked-language encoders; (ii) we add causal experiments that test whether aligned states drive the receiving model's behaviour (Section~\ref{sec:crossmodel_patching}); (iii) we vary training data over four controlled tiers (Section~\ref{sec:size_scaling}); and (iv) we replicate the finding across five ${\sim}$1B models developed by different research groups (Section~\ref{sec:cross_arch}).

\section{Representational Alignment}
\label{sec:representational_alignment}

We first ask whether independently trained monolingual models converge on similar internal representations of parallel semantic content. We assess this correlationally by measuring CKA between model pairs under matched and shuffled controls.

\subsection{Experimental Setup}
\label{sec:setup_cka}

\paragraph{Models.}
We start by using strictly monolingual Goldfish causal language models \citep{chang2024goldfish}.\footnote{\url{https://huggingface.co/goldfish-models}} All models share an identical Transformer architecture and training data size (adjusted for amount of information content), but have disjoint vocabularies, parameters, and training corpora. The family uses two architectures matched to data scale: the 5\,MB and 10\,MB models have 39M parameters, while the 100\,MB and 1000\,MB models have 125M parameters. Unless otherwise stated, we use the 1000\,MB tier.\footnote{A GlotLID-v3 audit of the training corpora (Appendix~\ref{sec:appendix_contamination}) finds at most 0.1\% English content in the non-English corpora used in the main analysis.}

\paragraph{Parallel data.}
We evaluate on four parallel corpora: FLORES-200 \citep{costa2022no}, Tatoeba \citep{tiedemann-2020-tatoeba}, OPUS \citep{TIEDEMANN12.463}, and BouQUET \citep{andrews-etal-2025-bouquet}. For each dataset and language pair, we extract hidden states from every Transformer layer, including the embedding layer.

\paragraph{Similarity metric.}
We measure cross-model similarity with linear CKA \citep{kornblith2019similarity}:
\[
\mathrm{CKA}(X, Y) =
\frac{\mathrm{HSIC}(X, Y)}{\sqrt{\mathrm{HSIC}(X, X)\,\mathrm{HSIC}(Y, Y)}},
\]
which is invariant to orthogonal transformations and isotropic rescaling. To control for spurious similarity, we contrast a \emph{matched} condition, which consists of aligned parallel sentences, with a \emph{shuffled} control. For a language pair $(A, B)$, the control keeps side $A$ in order and reorders the rows of side $B$ by a single random permutation (fixed seed), drawn once per pair and applied at every layer: each $A$ sentence is paired with a $B$ sentence that is not its translation, and every $B$ sentence appears exactly once. The control therefore preserves the architecture and corpus statistics of both sides, so the matched$-$shuffled gap isolates the semantic correspondence between paired sentences. We compare across all 36 language pairs.

\paragraph{Three representations.}
We compute CKA under three pooling strategies. Our sentence-level representation includes
(i) \emph{Mean pooling}: average token hidden states across the sequence; (ii) \emph{SGPT position-weighted pooling} \citep{muennighoff2022sgpt}: weight each token by its position in the sequence before averaging, emphasising information accumulated towards the end of the context, which proves to work better for causal language models; and 
(iii) \emph{Token-level word-aligned pooling}: isolate lexical correspondence by averaging subword hidden states within each aligned word span, using \textsc{SimAlign} \citep{jalili-sabet-etal-2020-simalign} word alignments over the parallel sentence pair. For a word $w$ with subword token set $T(w)$ and per-token hidden states $\mathbf{h}_{t,\ell}$ at layer $\ell$, $\mathbf{z}_{w,\ell} = \frac{1}{|T(w)|}\sum_{t\in T(w)} \mathbf{h}_{t,\ell}$.

Per-pair CKA values for all three settings on all four datasets are in Appendix~\ref{sec:appendix_cka}.

\subsection{Cross-Lingual Alignment Across 36 Pairs}
\label{sec:core_findings}

\begin{figure}[t]
\centering
\includegraphics[width=\linewidth]{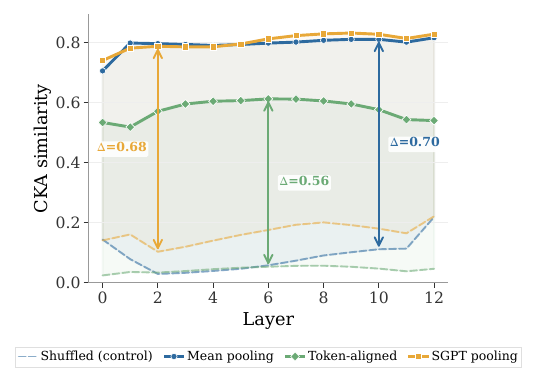}
\caption{\textbf{Layerwise English--French CKA} (Goldfish 1000\,MB) under three representations (mean pooling, token-aligned, SGPT). Solid: matched; dashed: shuffled.}

\label{fig:cka_layers_main}
\end{figure}

\begin{figure}[t]
\centering
\includegraphics[width=\linewidth]{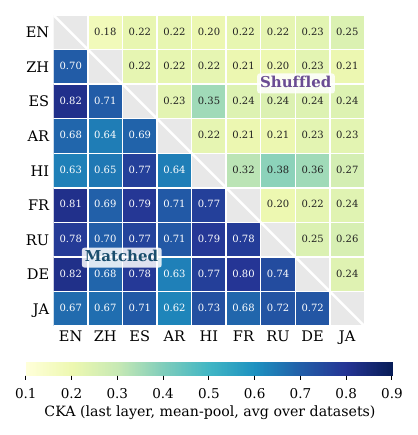}
\caption{\textbf{Last-layer pairwise CKA (Goldfish 1000\,MB, FLORES-200).} Lower triangle: matched; upper triangle: shuffled.}
\label{fig:cka_heatmap_main}
\end{figure}

We show the gap between the matched and shuffled conditions in  Figure~\ref{fig:cka_layers_main}. We show the per-layer trajectory for a single representative pair (English--French) under all three pooling-strategies and the shuffled condition. We also report $\Delta$ between the two, where a larger delta indicates more improvement over the baseline.
Table~\ref{tab:cka_summary} averages the gap across all pairs in four datasets.
Figure~\ref{fig:cka_heatmap_main} unpacks the averages to a $9\times 9$ matrix. 

\begin{table*}[t]
\centering
\small
\setlength{\tabcolsep}{5pt}
\renewcommand{\arraystretch}{1.1}
\begin{tabular}{l ccc c ccc c ccc}
\toprule
 & \multicolumn{3}{c}{\textbf{Mean Pooling}} & & \multicolumn{3}{c}{\textbf{Token-Aligned}} & & \multicolumn{3}{c}{\textbf{SGPT}} \\
\cmidrule{2-4} \cmidrule{6-8} \cmidrule{10-12}
\textbf{Dataset} & Matched & Shuffled & $\Delta$ & & Matched & Shuffled & $\Delta$ & & Matched & Shuffled & $\Delta$ \\
\midrule
FLORES   & 0.72 & 0.15 & +0.57 & & 0.38 & 0.03 & +0.35 & & \textbf{0.78} & 0.17 & +0.61 \\
Tatoeba  & 0.69 & 0.16 & +0.53 & & 0.36 & 0.03 & +0.33 & & \textbf{0.75} & 0.18 & +0.57 \\
OPUS     & 0.68 & 0.15 & +0.53 & & 0.35 & 0.03 & +0.32 & & \textbf{0.73} & 0.17 & +0.56 \\
BouQUET  & 0.70 & 0.16 & +0.54 & & 0.37 & 0.03 & +0.34 & & \textbf{0.76} & 0.18 & +0.58 \\
\midrule
\emph{Avg.} & \emph{0.70} & \emph{0.15} & \emph{+0.54} & & \emph{0.36} & \emph{0.03} & \emph{+0.34} & & \emph{0.76} & \emph{0.18} & \emph{+0.58} \\
\bottomrule
\end{tabular}
\caption{\textbf{Mean last-layer CKA across all 36 Goldfish (1000\,MB) language pairs}, by representation and dataset. $\Delta$ is the matched$-$shuffled gap. Full per-dataset $9\times 9$ matrices are in Appendix Tables~\ref{tab:cka_mean_pool_flores}--\ref{tab:cka_sgpt_bouquet}.}
\label{tab:cka_summary}
\end{table*}

\paragraph{Semantic alignment exceeds architectural noise.}
Matched CKA is higher than the shuffled baseline at every layer and for every language pair (Figures~\ref{fig:cka_layers_main} and~\ref{fig:cka_heatmap_main}). Under SGPT pooling, last-layer matched CKA averages 0.78 on FLORES, compared to 0.17 shuffled; the consistent gap is preserved across datasets (Table~\ref{tab:cka_summary}). Since the shuffled control uses the same sentences and only randomises the pairing, this gap cannot be explained by shared Transformer architecture or surface token statistics. Therefore, these results indicate stronger alignment than would be expected by chance.

\paragraph{Alignment is consistent across depth.}
Matched CKA is stable across layers under all three representations (Figure~\ref{fig:cka_layers_main}): in every case the layer-to-layer variation is small relative to the matched--shuffled gap. 

\paragraph{Linguistic similarity modulates alignment.}
Figure~\ref{fig:cka_heatmap_main} suggests a typological gradient: linguistically related pairs such as English--French (0.81), English--German (0.82) exhibit stronger alignment than distant pairs such as Chinese--Arabic (0.64) or Chinese--Hindi (0.65). We further test this by correlating last-layer matched CKA against URIEL typological distances \citep{littell-etal-2017-uriel} and a measure of model performance across the 36 Goldfish pairs. Following \citet{chang2024goldfish}, we use mean sentence negative log-likelihood (NLL) over parallel sentences to measure model quality. URIEL syntactic distance is the strongest single predictor ($\rho{=}{-}0.64$, $p{<}.001$). This is consistent with previous work that shows that more similar languages have higher representational overlap and more cross-lingual transfer \citep{chang-etal-2024-multilinguality, arnett-etal-2025-acquisition}. Mean NLL is also a significant predictor of CKA alignment ($\rho{=}{-}0.43$, $p{=}.01$), which suggests that the quality of the model's representations affects cross-lingual alignability: better models are more aligned across languages. This is predicted by the Platonic Representation Hypothesis, which states that representational convergence increases with model capability. A joint regression confirms that syntactic distance is the dominant predictor, improving fit over mean NLL alone ($\Delta R^2{=}.19$, $F(1,33){=}9.88$, $p{=}.004$). Full correlations and visualisations are provided in Appendix~\ref{sec:appendix_explaining_alignment}.

\paragraph{Low-resource languages also confirm alignment.}
The analyses thus far have been limited to high-resource languages. We test whether these results hold across different language families and writing systems.
We repeat the same evaluation with the English model paired against four real-world low-resource languages: Tagalog (\texttt{tgl}), Swahili (\texttt{swh}), Northern Uzbek (\texttt{uzn}), and Amharic (\texttt{amh}). Table~\ref{tab:cka_lowresource} reports matched and shuffled CKA at the last and middle layers: the gaps (${+}0.39$ to ${+}0.58$) sit in the same range as the high-resource reference pairs, and hold at the middle layer as well.

\begin{table}[t]
\centering
\small
\setlength{\tabcolsep}{3pt}
\renewcommand{\arraystretch}{1.1}
\resizebox{\ifdim\width>\columnwidth\columnwidth\else\width\fi}{!}{%
\begin{tabular}{ll cc}
\toprule
 & & \textbf{Last layer} & \textbf{Middle layer} \\
\textbf{Pair} & \textbf{Family (script)} & M\,/\,S\,($\Delta$) & M\,/\,S\,($\Delta$) \\
\midrule
eng--tgl & Austrones.\ (Latn)  & .83\,/\,.25\,(+.58) & .84\,/\,.25\,(+.59) \\
eng--swh & Atl.--Congo (Latn) & .81\,/\,.25\,(+.56) & .82\,/\,.24\,(+.59) \\
eng--uzn & Turkic (Latn)       & .80\,/\,.26\,(+.54) & .80\,/\,.25\,(+.55) \\
eng--amh & Afro-As.\ (Ethi)    & .80\,/\,.41\,(+.39) & .81\,/\,.41\,(+.41) \\
\midrule
eng--fra & Indo-Eur.\ (Latn)   & .86\,/\,.24\,(+.62) & --- \\
eng--zho & Sino-Tib.\ (Hans)   & .82\,/\,.23\,(+.59) & --- \\
\bottomrule
\end{tabular}}
\caption{Matched (M) and shuffled (S) SGPT CKA and their gap ($\Delta$) for English paired with four low-resource languages, at the last and middle ($L/2$) layer. Goldfish 1000\,MB models, FLORES-200, conditions as in Section~\ref{sec:setup_cka}. Reference values for eng--fra and eng--zho are the last-layer results from Table~\ref{tab:cka_sgpt_flores}.}
\label{tab:cka_lowresource}
\end{table}

\subsection{Alignment Scales with Training Data}
\label{sec:size_scaling}

\begin{figure}[t]
\centering
\includegraphics[width=\linewidth]{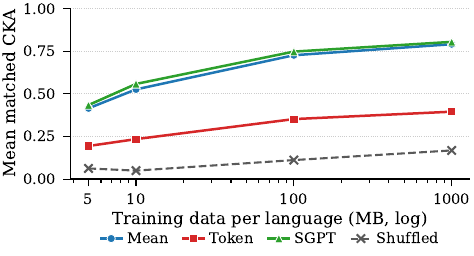}
\caption{\textbf{Alignment scales with training data.} Mean matched last-layer CKA grows monotonically from 5\,MB to 1000\,MB under all three representations. Full $9\times9$ heatmaps at every scale are provided in Appendix Figures~\ref{fig:size_heatmap_mean}--\ref{fig:size_heatmap_sgpt}.}
\label{fig:size_scaling}
\end{figure}

We use all four training-data scales available for Goldfish (5, 10, 100, 1000\,MB) for each language, enabling a controlled study of how alignment grows with training data. Figure~\ref{fig:size_scaling} plots mean off-diagonal matched CKA against training size for each of the three representations, with shuffled baselines for reference. Across all three representations, matched CKA increases monotonically with training-data scale. 

A cross-size comparison pairs Goldfish models trained on different amounts of data (1000\,MB against 100\,MB, and 10\,MB against 5\,MB) within each model size. We see the diagonal (same language) dominates every row, and off-diagonal matched CKA still sits above the shuffled baseline at both pairings. This supports the conclusions that as model quality improves, cross-lingual representational alignment gets stronger, which is consistent with both the Platonic Representation Hypothesis and our results in Section \ref{sec:core_findings}.  We provide further details about this analysis in Appendix~\ref{sec:appendix_size_cross}.

\subsection{Alignment Persists Between Larger Models}
\label{sec:cross_arch}

A natural concern is that alignment within Goldfish models is only an artefact from shared design choices across the family. We therefore test whether alignment survives when models differ in architecture, vocabulary, training data, and research group. As a first check, comparing Goldfish English against Pythia-160M~\citep{biderman2023pythia} and Goldfish Chinese against Zh-Pythia-160M~\citep{liu2025systematicassessmentlanguagemodels} also yields matched SGPT CKA above the shuffled baseline at every layer, albeit in a U-shaped profile (more discussion in Appendix~\ref{sec:appendix_cross_arch_pythia}). We then compare five independently developed $\sim$1B monolingual models: Pythia-1.4B EN, Zh-Pythia-1.4B, Tucano-1B PT~\citep{correa2025tucanoadvancingneuraltext}, Bielik-1.5B PL~\citep{ociepa2025bielikv3smalltechnical}, Minerva-1B IT~\citep{orlando-etal-2024-minerva}, full specifications in Appendix Table~\ref{tab:diverse_models}. We conduct the same experiment as above. Table~\ref{tab:diverse_main} reports last-layer CKA (SGPT) for all $\binom{5}{2}=10$ model pairs on FLORES-200. Matched CKA exceeds the shuffled baseline in all 10 pairs (0.71 vs.\ 0.18 on average), confirming that the alignment pattern extends to models sharing nothing but the objective of modelling text.

\begin{table}[t]
\centering
\small
\setlength{\tabcolsep}{3pt}
\renewcommand{\arraystretch}{1.1}
\begin{tabular}{l|ccccc}
 & \rotatebox{55}{\textbf{Pyth.\,EN}} & \rotatebox{55}{\textbf{Pyth.\,ZH}} & \rotatebox{55}{\textbf{Tuc.\,PT}} & \rotatebox{55}{\textbf{Biel.\,PL}} & \rotatebox{55}{\textbf{Min.\,IT}} \\
\hline
\textbf{Pyth.\,EN}    & \cellcolor{gray!20}---   & \cellcolor{red!8}.16  & \cellcolor{red!10}.19  & \cellcolor{red!6}.13  & \cellcolor{red!12}.22 \\
\textbf{Pyth.\,ZH}  & \cellcolor{blue!25}.70   & \cellcolor{gray!20}--- & \cellcolor{red!9}.18  & \cellcolor{red!5}.12  & \cellcolor{red!11}.21 \\
\textbf{Tuc.\,PT}     & \cellcolor{blue!50}.84   & \cellcolor{blue!23}.69 & \cellcolor{gray!20}--- & \cellcolor{red!7}.14  & \cellcolor{red!13}.25 \\
\textbf{Biel.\,PL}     & \cellcolor{blue!20}.67   & \cellcolor{blue!15}.61 & \cellcolor{blue!19}.66 & \cellcolor{gray!20}--- & \cellcolor{red!9}.17 \\
\textbf{Min.\,IT}    & \cellcolor{blue!45}.82   & \cellcolor{blue!25}.70 & \cellcolor{blue!43}.80 & \cellcolor{blue!17}.63 & \cellcolor{gray!20}--- \\
\end{tabular}
\caption{\textbf{Last-layer CKA across five $\sim$1B monolingual models} (SGPT). Lower triangle (blue): matched. Upper triangle (red): shuffled. Detailed model specifications are in Appendix Table~\ref{tab:diverse_models}.}
\label{tab:diverse_main}
\end{table}

CKA establishes similarity up to an unspecified transformation: it shows the two spaces are alignable but does not identify the map that aligns them.
In the following section, we will establish causal evidence for meaningful alignment across models.

\section{Cross-Lingual Representation Reconstruction}
\label{sec:reconstruction}

Section~\ref{sec:representational_alignment} established that independently trained monolingual models produce representations whose pairwise similarity exceeds shuffled baselines. We now move from correlation to \emph{reconstruction},\footnote{We use \emph{construction} and \emph{reconstruction} interchangeably for these experiments.} asking what class of learned map recovers one model's representations from another's.

\subsection{Setup}
\label{sec:reconstruction_setup}

We study all pairs of the Goldfish 1000\,MB monolingual models from Section~\ref{sec:representational_alignment}, and fit projections in both directions, yielding 72 directional pairs in total. As SGPT resulted in the strongest alignment, we use this method at every layer including the embedding layer. We use the same four parallel datasets as Section~\ref{sec:representational_alignment}, yielding $\sim$5{,}000 parallel sentences per language pair (80/20 train/test split, seed 42).\footnote{$\leq$50-character sentences filtered out.} On top of the resulting per-layer matrices $X \in \mathbb{R}^{n \times d}$ (source) and $Y \in \mathbb{R}^{n \times d}$ (target), we fit a hierarchy of three projection methods that each relax a structural constraint. All three minimise squared reconstruction error.

\paragraph{(i) Procrustes.}
The most constrained mapping: a single rotation/reflection $W \in \mathbb{R}^{d\times d}$ that preserves all pairwise inner products in the source space \citep{schonemann1966generalized}.
\begin{equation}
W^{\star} = \operatorname*{arg\,min}_{W^\top W = I}\;\|XW - Y\|_{F}^{2},
\end{equation}
solved in closed form by $W^{\star}=UV^{\top}$, where $U\Sigma V^{\top}=X^{\top}Y$ is the SVD.

\paragraph{(ii) Affine.}
We relax orthogonality: $W$ is unconstrained and a bias $b\in\mathbb{R}^{d}$ is allowed.\footnote{Solved in closed form via centring with ridge regularisation $\lambda\|W\|_{F}^{2}$; $\lambda$ is tuned on a held-out validation slice over $\{10^{-2},\dots,10^{4}\}$.}
\begin{equation}
(W^{\star},b^{\star}) = \operatorname*{arg\,min}_{W,b}\;\|XW + \mathbf{1}b^{\top} - Y\|_{F}^{2}.
\end{equation}

\paragraph{(iii) MLP.}
We further relax linearity using a one-hidden-layer ReLU network $f_{\theta}$ of width $d$.\footnote{Trained with Adam ($\mathrm{lr}{=}10^{-4}$), weight decay $10^{-2}$, batch size $64$, up to $200$ epochs with early stopping (patience $20$) on a 10\% internal validation split.}
\begin{align}
\theta^{\star} &= \operatorname*{arg\,min}_{\theta}\;\|f_{\theta}(X) - Y\|_{F}^{2}, \\
f_{\theta}(x) &= W_{2}\,\mathrm{ReLU}(W_{1}x+b_{1})+b_{2}.
\end{align}

\smallskip
We evaluate on three complementary metrics with distinct geometric meaning: \textbf{P@1} (cosine retrieval, capturing local neighbourhood preservation), \textbf{MSE} (mean squared error; pointwise reconstruction quality), and post-projection \textbf{linear CKA} (relational structure). P@1 is reported in two variants: P@1${_\text{std}}$ retrieves against the test pool, while P@1${_\text{hard}}$ retrieves against the combined train${+}$test pool. 

\subsection{Rotation Is Nearly Sufficient}
\label{sec:reconstruction_core}

\begin{table}[t]
\centering
\small
\setlength{\tabcolsep}{4pt}
\renewcommand{\arraystretch}{1}
\resizebox{\ifdim\width>\columnwidth\columnwidth\else\width\fi}{!}{%
\begin{tabular}{l cccc}
\toprule
\textbf{Method} & P@1$_{\text{std}}$ & P@1$_{\text{hard}}$ & MSE & CKA \\
\midrule
Procrustes & $\mathbf{.887}_{\pm.021}$ & $\mathbf{.775}_{\pm.028}$ & $.689_{\pm.026}$ & $.713_{\pm.023}$ \\
Affine     & $.814_{\pm.027}$ & $.663_{\pm.036}$ & $\mathbf{.439}_{\pm.017}$ & $\mathbf{.762}_{\pm.020}$ \\
MLP        & $.809_{\pm.029}$ & $.643_{\pm.038}$ & $.456_{\pm.017}$ & $.760_{\pm.020}$ \\
\midrule
Identity   & .002 & .000 & 2.258 & .713 \\
\bottomrule
\end{tabular}}
\caption{\textbf{Method comparison across 36 Goldfish language pairs} (final layer). Each cell shows the mean across 36 pairs with bootstrapped 95\% CI (10{,}000 resamples); \textbf{bold} marks the best method per metric. P@1$_{\text{std}}$ retrieves against the test-only pool; P@1$_{\text{hard}}$ against the combined train${+}$test pool. MSE and CKA are pool-independent. Per-pair scatter is in Appendix Figures~\ref{fig:procrustes_vs_affine} and~\ref{fig:methods_x_metrics}.}
\label{tab:reconstruction_summary}
\end{table}

Table~\ref{tab:reconstruction_summary} summarises reconstruction across all 72 directions. Three observations follow. 
\textbf{(i) All three projections are highly effective.} The identity baseline retrieves nothing (P@1$_{\text{std}}{=}0.2\%$); a single Procrustes rotation leads to  $88.7\%$ retrieval accuracy, with MSE down from $2.258$ to $0.689$. A learned mapping recovers the matching target representation in the vast majority of cases. \textbf{(ii) Affine and MLP win MSE and CKA, as expected.} Both methods minimise reconstruction error directly and operate in a strictly larger hypothesis class than Procrustes, so unsurprisingly they attain lower MSE on all 36 pairs (Affine $0.439$, MLP $0.456$ vs.\ Procrustes $0.689$) and higher post-projection CKA (rotation does not change CKA).
\textbf{(iii) Procrustes nonetheless wins retrieval.} Procrustes beats the more flexible methods on P@1 on both the standard and the hard pool, while its MSE stays within range of Affine's. We hypothesise this as an objective-metric mismatch: every method minimises Euclidean reconstruction error, but P@1 is rank-based and direction-sensitive. Because Procrustes is pure rotation, it preserves every angle in the source space, whereas Affine and MLP have no such constraint and would rescale axes anisotropically, lowering MSE at the cost of preserving angular geometry. The orthogonality constraint preserves better neighbourhood structure.

\begin{figure}[t]
    \centering
    \includegraphics[width=\linewidth]{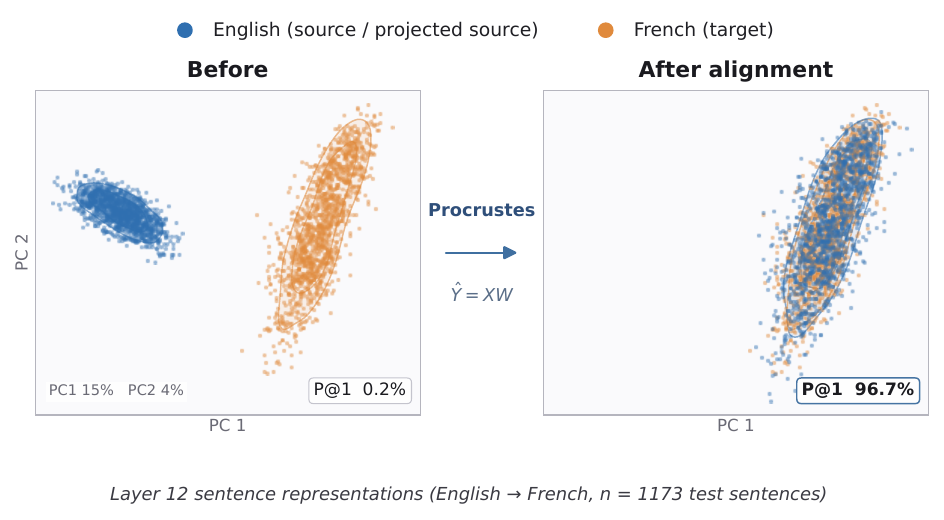}
    \caption{\textbf{Procrustes mapping, eng--fra, layer 12} (2D PCA over the joint test set). English in blue (source, Procrustes-mapped); French in yellow (target). The per-layer full version is in Figure~\ref{fig:pca_procrustes_engfra}.}
    \label{fig:pca_procrustes_main}
\end{figure}

Figure~\ref{fig:pca_procrustes_main} visualises this directly for \texttt{eng}--\texttt{fra} at the final layer: projecting both languages into a shared 2D PCA basis and overlaying the Procrustes-mapped English representations on the matched French targets, the two clouds coincide while the mapped source retains the geometry of the original English cloud. This is related to a pre-Transformer line of work on word-embedding alignment, where orthogonal mappings are shown to outperform unconstrained linear ones on retrieval \citep{mikolov2013exploiting,xing-etal-2015-normalized,smith2017offline,artetxe-etal-2018-robust,conneau2018word}.  We observe the same dissociation in transformer hidden states.

\subsection{Layers and residual structure analysis}

Next, we verify that the mapping we find is reliable across layers, not only at the final layer reported in Table~\ref{tab:reconstruction_summary}. Sweeping Procrustes across all 13 layers (Appendix~\ref{sec:reconstruction_layers}), retrieval stays well above the identity baseline at every layer and peaks in mid layers (6--8): \texttt{eng}--\texttt{fra} P@1${=}0.981$ at layer 8; \texttt{eng}--\texttt{rus} $0.987$ at layer 6. We confirm that rotation is usable across the depth of the network. A complementary dimensional analysis (Appendix~\ref{sec:reconstruction_dim}) examines where the post-Procrustes residual lives. The residual dimension is high-rank and concentrated outside the principal directions of the target; this is why neither Affine relaxation nor a one-layer MLP improves retrieval.

Our reconstruction experiment so far works geometrically. In Section~\ref{sec:crossmodel_patching} below we test for causal evidence that this rotation results in the best downstream result.

\section{Cross-Model Activation Patching}
\label{sec:crossmodel_patching}

Reconstruction shows that a single rotation maps one model's representations onto another's geometrically. We now ask whether the same rotation, applied with no further training, could transfer to a different task functionally. We test this on a factual recall task using cross-model activation patching, which substitutes a projected residual into another model's forward pass and measures whether the prediction follows the substitution.

\subsection{Task and Intervention}
\label{sec:patching_setup}

\paragraph{Factual probe.}
We construct country$\to$capital facts in six languages (English, French, German, Spanish, Japanese, Chinese), with hand-curated cloze prompts and answers, e.g.\ \texttt{``The capital of France is''} $\to$ \texttt{Paris}; \texttt{``Die Hauptstadt von Frankreich ist''} $\to$ \texttt{Paris}. The country word's last subword position is identified per language by a verb-marker heuristic (details in Appendix~\ref{app:patching}).

\paragraph{Span patching at the concept position.} For a (donor, target) cross-fact pair (with donor $\neq$ target country), we run the source-language model on the donor's source-language prompt and cache the residual stream $\mathbf{h}_S^{(k)}[c_S]$ at the donor country's last subword position $c_S$, for every layer $k\in\{0,\dots,L\}$ ($L{=}12$). The target-language model is then run on its own target-language target prompt, but at the target country position $c_T$, we replace the residual $\mathbf{h}_T^{(k)}[c_T]$ with the projected donor residual $f^{(k)}\!\left(\mathbf{h}_S^{(k)}[c_S]\right)$ for all $k\in\{j,\dots,L\}$. The intervention follows two design choices from \citet{dumas-etal-2025-separating}: (i) the patch spans layers $j\!\to\!L$ rather than firing at a single layer, since a single-layer last-token patch is suppressed by downstream computation; and (ii) the patch lands on the concept word rather than the last token, so that the injected information must propagate through attention to influence the answer logits.

\paragraph{Conditions.}
We compare four sources for the patched residual: \textit{within-lang} ($\mathbf{h}_T$, the target model patches its own donor representation, no projection; the ceiling on what any cross-lingual map could match), \textit{Procrustes} ($W^{(k)}\mathbf{h}_S$, the rotation described in Section~\ref{sec:reconstruction}, applied per layer with no further training); \textit{unprojected} ($\mathbf{h}_S$, raw source-model residual, basis-mismatch control); and \textit{shuffled} ($\sim \mathcal{N}(\boldsymbol\mu_S, \boldsymbol\Sigma_S)$, a Gaussian sample per layer matching the source model's activation mean and full covariance, distributional control). Affine (Section~\ref{sec:reconstruction}) was also evaluated but is excluded since it underperforms Procrustes on this probe; see Appendix~\ref{app:patching} for the explanation.

\paragraph{Metric.}
We score with the \emph{directional success rate}: the fraction of cross-fact pairs where $\Delta\log p(\text{donor answer}) > 0$ in the target language under patching, with Wilson 95\% CIs over $n$ cross-fact pairs per cell ($n{=}870$ for country$\to$capital). The directional rate is approximately invariant to source-model strength asymmetries; raw $\Delta\log p$ is contaminated by the fact that some models hold country$\to$capital knowledge more strongly than others.

\subsection{Cross-Model Transfer of Factual Content}
\label{sec:patching_results}

\begin{figure}[t]
\centering
\includegraphics[width=\columnwidth]{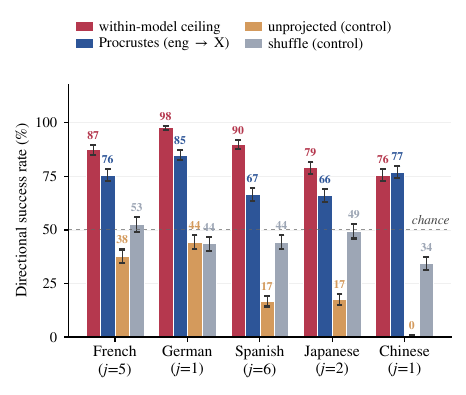}
\caption{\textbf{Directional success rate (\%) for \texttt{eng}${\to}$\{\texttt{fra}, \texttt{deu}, \texttt{spa}, \texttt{jpn}, \texttt{zho}\}}, $n{=}870$ cross-fact pairs per cell, Wilson 95\% CIs. Bars: within-model ceiling, Procrustes, unprojected residual, and shuffled Gaussian sample. Per target, the patch start layer $j\in\{1,\dots,8\}$ is selected to maximise \texttt{within} and \texttt{Procrustes}. Dashed line at 50\% chance.}
\label{fig:patching_summary}
\end{figure}

We report Procrustes as the cross-lingual projection of record (see Appendix~\ref{app:patching} for discussions on Affine). Figure~\ref{fig:patching_summary} reports cross-model transfer across five \texttt{eng}$\to$\{\texttt{fra}, \texttt{deu}, \texttt{spa}, \texttt{jpn}, \texttt{zho}\} target languages at the patch-start layer that best separates Procrustes from controls per target. Four observations follow.

\textbf{(i) Establishing the ceiling by patching with the same-language representation.} Before evaluating any cross-model projection we verify that the intervention itself can carry factual content when no cross-lingual transfer is required. Within-model patching yields directional success rates of $76$--$98\%$ across the five targets, reproducing the \citet{dumas-etal-2025-separating} concept-position span-patch design in our setting and giving an upper bound on what any cross-lingual map could match.

\textbf{(ii) The fitted rotation from Section~\ref{sec:reconstruction} successfully transfers factual concepts between monolingual models.} The Procrustes map from Section~\ref{sec:reconstruction} was fit on sentence-end pooled activations. We evaluate it here, without retraining, on a different task (factual cloze rather than news) at a different token position (the concept word mid-prompt rather than the sequence end). We score against a different readout (next-token logits rather than cosine retrieval). Remarkably, the rotation reaches directional success rates of $66$--$85\%$ across the five targets, against a $76$--$98\%$ within-model ceiling, recovering most of what an oracle same-model intervention would achieve. We also see that the pattern holds across different languages and scripts.

\textbf{(iii) Control settings (shuffle and unprojected) are cleanly near or below the chance line.} The shuffled control sits approximately at the $50\%$ chance line, as expected for a random perturbation. However, the unprojected control falls clearly below chance. We hypothesise that unprojected carries the donor concept in a basis that is mismatched with the target model, so when read by the target model's LM head it acts as a coherent signal pointing in the wrong direction, and sometimes produces the opposite effect.

\textbf{(iv) The rotation transfers across factual relations.} Appendix Table~\ref{tab:patching_relations} extends the probe from country$\to$capital to six further relations between the English and German models, reusing the same per-layer Procrustes maps with no refitting and a fixed patch start $j{=}2$. Several relations involve naturally multi-token entities (e.g.\ work titles spanning several words on the prompt side); all conditions score the answer at its first subword. Procrustes exceeds the shuffled control on all seven relations; the Wilson intervals separate on the four larger sets ($n \geq 398$), and the remaining three are directionally consistent. We read the patching experiments as a proof of existence: a rotation fitted once on parallel sentence pools transfers zero-shot to a new task, a new token position, and new relations. Breadth of deployment, multi-token generation, and open-ended use remain outside this design (see Limitations).

Taking \texttt{eng}$\to$\texttt{deu} as a concrete illustration: patching the German model's own donor representation flips the prediction to the donor's capital $98\%$ of the time; the rotated English residual achieves $85\%$ of cross-fact pairs; a random Gaussian sample sits near $50\%$. A single orthogonal map, fit once on parallel sentence pools, recovers most of what an oracle same-model intervention would do, while uninformed signals stay at or below chance.

\section{Conclusion and Discussion}

We set out asking whether strictly monolingual language models could be aligned across languages. In the \textbf{correlational} experiments, alignment between models exceeds shuffled controls on all pairs. In the \textbf{constructional} experiments, an orthogonal rotation recovers one model's hidden states from another's. In the \textbf{causal} experiments, the same rotation could patch an English residual into a German model and flip the factual prediction in $85\%$ of cases. This evidence suggests that cross-lingual alignment can emerge from the structure of language and the world it describes, with no joint training required.

More broadly speaking, we could think of multilinguality as a special case of the more general phenomenon of representational convergence \citep{moschella2023relativerepresentationsenablezeroshot}. Language-agnostic shared representations observed in multilingual language models may be partly a result of joint multilingual pretraining and partly inherited from crosslingual representational convergence \citep{k2020crosslingualabilitymultilingualbert, artetxe-etal-2020-cross}. Our results suggest that the latter is substantial, and that alignment already present in monolingual models can be recovered \textit{post hoc}. 

This opens a modular alternative to current multilingual pipelines: a small set of anchor pairs is in principle enough to stitch independently trained monolingual specialists into a shared space \citep{lenc2015understanding, bansal2021revisiting}, and likewise for model merging across languages \citep{ilharco2023editing, yadav2023tiesmerging}. In preliminary follow-up experiments, we stitch two monolingual Goldfish models into a functioning bilingual system that approaches the performance of a jointly trained bilingual baseline \citep{arnett-etal-2025-acquisition}; a systematic treatment is future work. 

A further direction is to make alignment an explicit training objective, so that a low-resource monolingual model is trained to be alignable with a high-resource anchor and the \textit{post hoc} rotation studied here becomes a channel for cross-lingual transfer. How to scale anchor-based projection to typologically distant pairs remains open.

\section*{Limitations}

\paragraph{Language and model coverage.} The nine-language Goldfish analysis and the five-model $\sim$1B replication are weighted towards higher-resource, predominantly Indo-European languages, reflecting which truly monolingual checkpoints are publicly available. The low-resource extension (Table~\ref{tab:cka_lowresource}) covers the correlational experiments only; reconstruction and patching for those languages remain untested. We cap at $\sim$1B parameters because most larger releases marketed as monolingual retain measurable multilingual content in their pretraining data. Since alignment increases monotonically with training data (Section~\ref{sec:size_scaling}), we expect the same pattern to extend to 7B+ monolingual checkpoints, but this remains untested.

\paragraph{Language contamination.} Foreign-language text in nominally monolingual corpora can itself induce cross-lingual ability \citep{blevins-zettlemoyer-2022-language}. Our audit (Appendix~\ref{sec:appendix_contamination}) bounds English contamination at or below 0.1\% for the nine main-analysis corpora, too little to account for the observed alignment, but residual contamination remains a small unquantified contributor, particularly for the noisier low-resource corpora.

\paragraph{Parallel-data assumption.} Rotations are fitted on parallel sentence pairs. Although this is small in absolute terms, sentence-aligned data is still not free for low-resource languages, and we do not evaluate unsupervised mapping alternatives.

\paragraph{Scope of the causal evaluation.} Cross-model activation patching covers seven factual relations at a single concept-bearing span, with English--German for the relation sweep (Table~\ref{tab:patching_relations}) and country$\to$capital for the five-target language sweep (Figure~\ref{fig:patching_summary}). Scoring is restricted to the answer's first subword. The evaluation establishes that the fitted rotation carries functional content across tasks and relations; multi-token generation and open-ended use fall outside this design.

\section*{Ethical Considerations}
All models and corpora used in this paper are publicly released research artefacts, and we release no new model or corpus. The factual probes we author cover encyclopaedic relations over public entities and contain no personal data. The modular use we sketch in the Discussion, stitching or merging independently trained monolingual models, would also carry each component model's biases and failure modes into the combined system, so any such system needs its own evaluation.

\section*{Acknowledgements}
We acknowledge the support of the UKRI Frontier Research Grant EP/Y031350/1 (EQUATE). Suchir Salhan is funded by Cambridge University Press \& Assessment.  Some experiments were performed using resources provided by the Cambridge Service for Data Driven Discovery (CSD3) operated by the University of Cambridge Research Computing Service, provided by Dell EMC and Intel using Tier-2 funding from the Engineering and Physical Sciences Research Council (capital grant EP/T022159/1), and DiRAC funding from the Science and Technology Facilities Council.

\section*{Use of AI assistants.} We use AI coding assistants for implementation support. All experimental design, analyses, and claims are the authors' own.

\bibliography{custom}

\clearpage
\newpage

\appendix

\section{Full Pairwise CKA Results}
\label{sec:appendix_cka}

Tables~\ref{tab:cka_mean_pool_flores}--\ref{tab:cka_sgpt_bouquet} (Appendix~\ref{sec:appendix_floats}) report the complete pairwise last-layer linear CKA values for all $\binom{9}{2} = 36$ language pairs across nine monolingual 1000\,MB Goldfish models. Each cell shows \textit{matched\,/\,shuffled} CKA, where \textit{matched} denotes CKA computed on parallel (translation-equivalent) sentences and \textit{shuffled} denotes a control in which the sentence order of one language is randomly permuted. Tables are grouped by pooling method: sentence-level mean pooling (Tables~\ref{tab:cka_mean_pool_flores}--\ref{tab:cka_mean_pool_bouquet}), token-level word-aligned (Tables~\ref{tab:cka_token_aligned_flores}--\ref{tab:cka_token_aligned_bouquet}), and SGPT position-weighted pooling (Tables~\ref{tab:cka_sgpt_flores}--\ref{tab:cka_sgpt_bouquet}), each evaluated on FLORES, Tatoeba, OPUS, and BouQUET. Missing entries (---) indicate language pairs for which parallel data was unavailable in that dataset. Results for English paired with four low-resource languages (Tagalog, Swahili, Northern Uzbek, Amharic) are reported in Table~\ref{tab:cka_lowresource} (Section~\ref{sec:core_findings}).

\section{CKA Comparison with Training Data}
\label{sec:appendix_model_size}

This appendix supports Section~\ref{sec:size_scaling} with the full $9\times 9$ pairwise CKA matrices at each of the four Goldfish training-data scales (5\,MB, 10\,MB, 100\,MB, 1000\,MB), and with a cross-size comparison that pairs models trained on \emph{different} amounts of data. All analyses use last-layer pairwise linear CKA across the same nine monolingual Goldfish languages from Appendix~\ref{sec:appendix_cka}, with a shuffled control throughout.

\subsection{Does Alignment Grow with Training Data?}
\label{sec:appendix_size_within}

\begin{figure*}[!htbp]
    \centering
    \includegraphics[width=\linewidth]{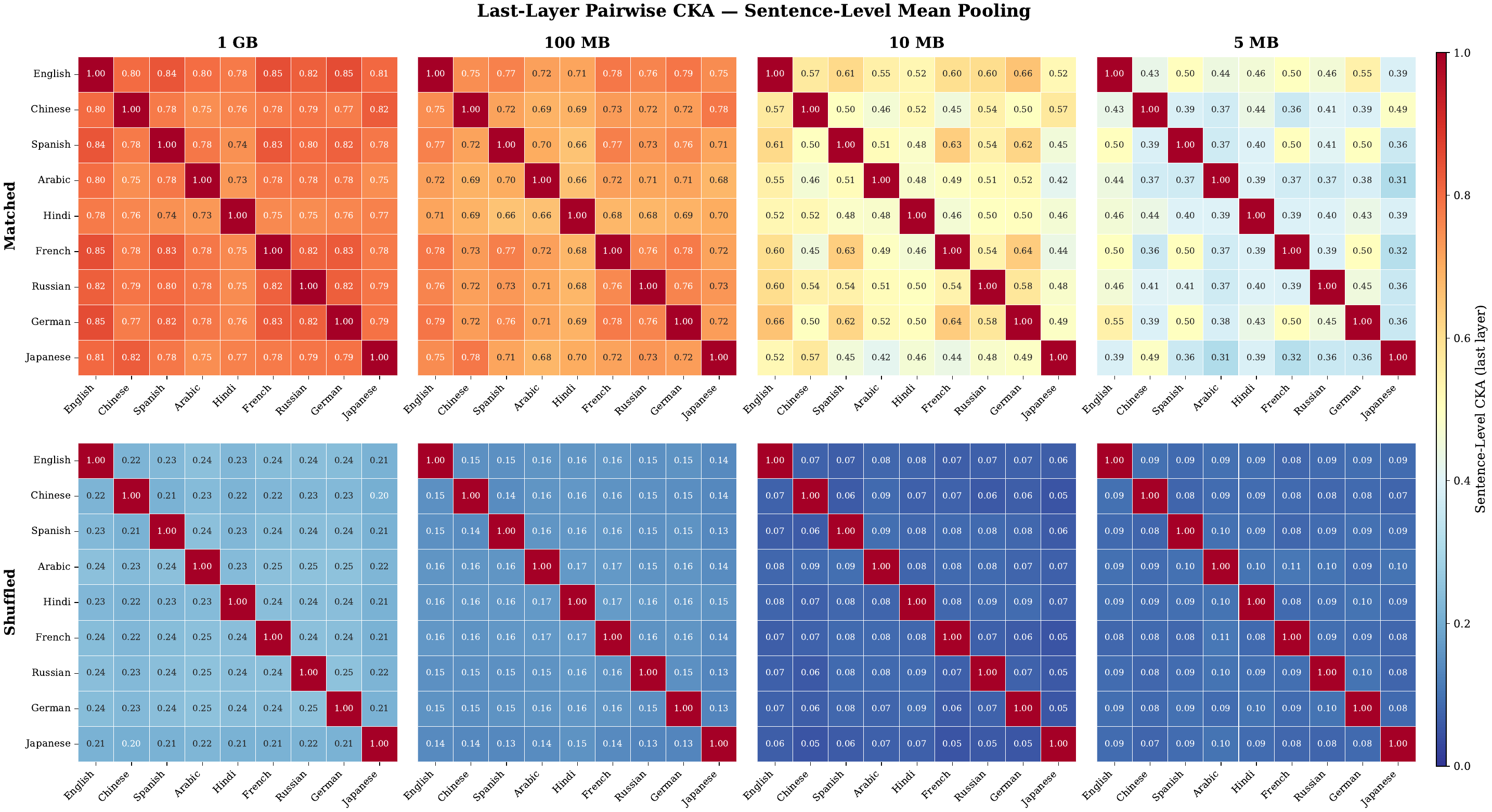}
    \caption{Setting 1 --- Sentence-Level Mean Pooling: last-layer pairwise CKA across all nine Goldfish languages at four training-data scales (5\,MB, 10\,MB, 100\,MB, 1000\,MB). Top row: matched CKA. Bottom row: shuffled baseline. Off-diagonal matched values rise monotonically with scale, while shuffled values remain near zero throughout.}
    \label{fig:size_heatmap_mean}
\end{figure*}

\begin{figure*}[!htbp]
    \centering
    \includegraphics[width=\linewidth]{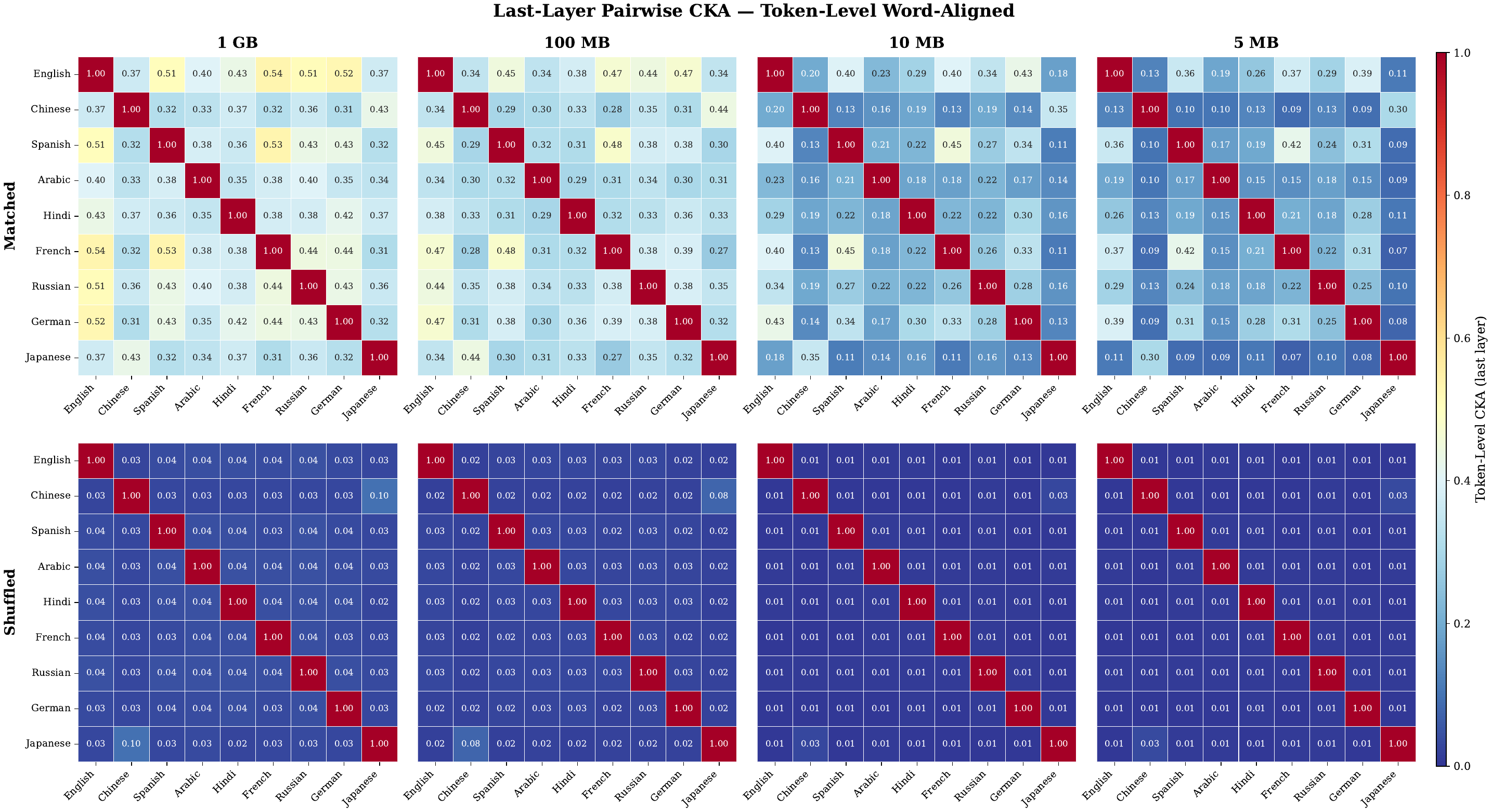}
    \caption{Setting 2 --- Token-Level Word-Aligned: same layout as Figure~\ref{fig:size_heatmap_mean}. Token-level alignment is lower in absolute terms than sentence-level but shows the same monotonic scaling with training data.}
    \label{fig:size_heatmap_token}
\end{figure*}

\begin{figure*}[!htbp]
    \centering
    \includegraphics[width=\linewidth]{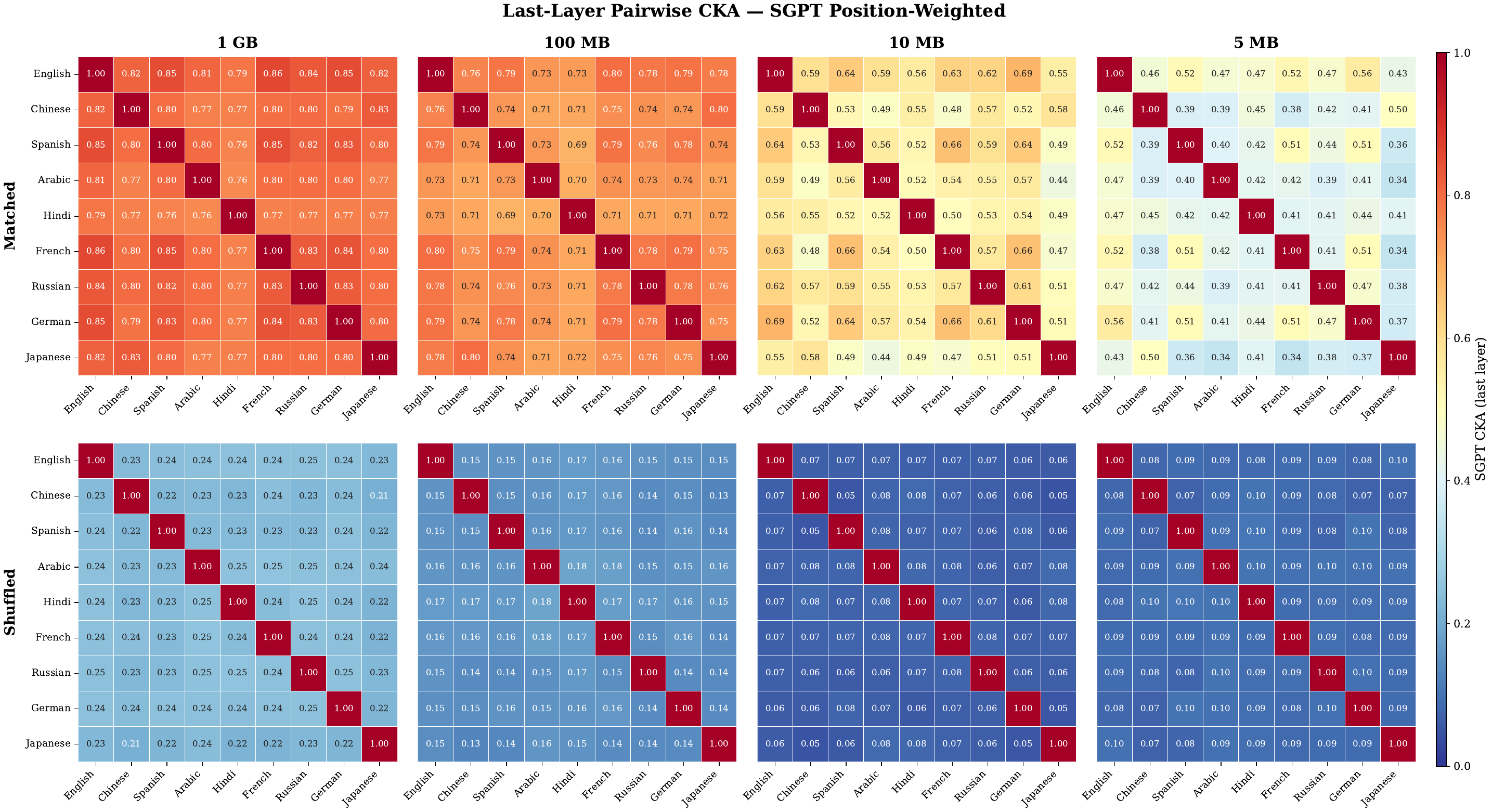}
    \caption{Setting 3 --- SGPT Position-Weighted Pooling: same layout as Figure~\ref{fig:size_heatmap_mean}. SGPT pooling yields the highest absolute matched values across all scales, while preserving the monotonic size effect.}
    \label{fig:size_heatmap_sgpt}
\end{figure*}

Figures~\ref{fig:size_heatmap_mean}--\ref{fig:size_heatmap_sgpt} (Appendix~\ref{sec:appendix_floats}) show the full $9\times9$ pairwise CKA matrices at each of the four data scales, for all three CKA settings: sentence-level mean pooling, token-level word-aligned, and SGPT position-weighted pooling.

\paragraph{Findings.}
Across all three CKA settings, matched off-diagonal CKA increases monotonically from 5\,MB to 1000\,MB. Shuffled baselines remain near zero at every scale, so the matched$-$shuffled gap widens with more training data. This confirms and extends the size-scaling result from Section~\ref{sec:size_scaling}: increased data exposure yields more semantically structured internal representations, consistently across all pooling strategies.

\FloatBarrier

\subsection{Cross-Size Comparison: Do Models of Different Scales Align?}
\label{sec:appendix_size_cross}

\begin{figure}[h]
    \centering
    \includegraphics[width=\linewidth]{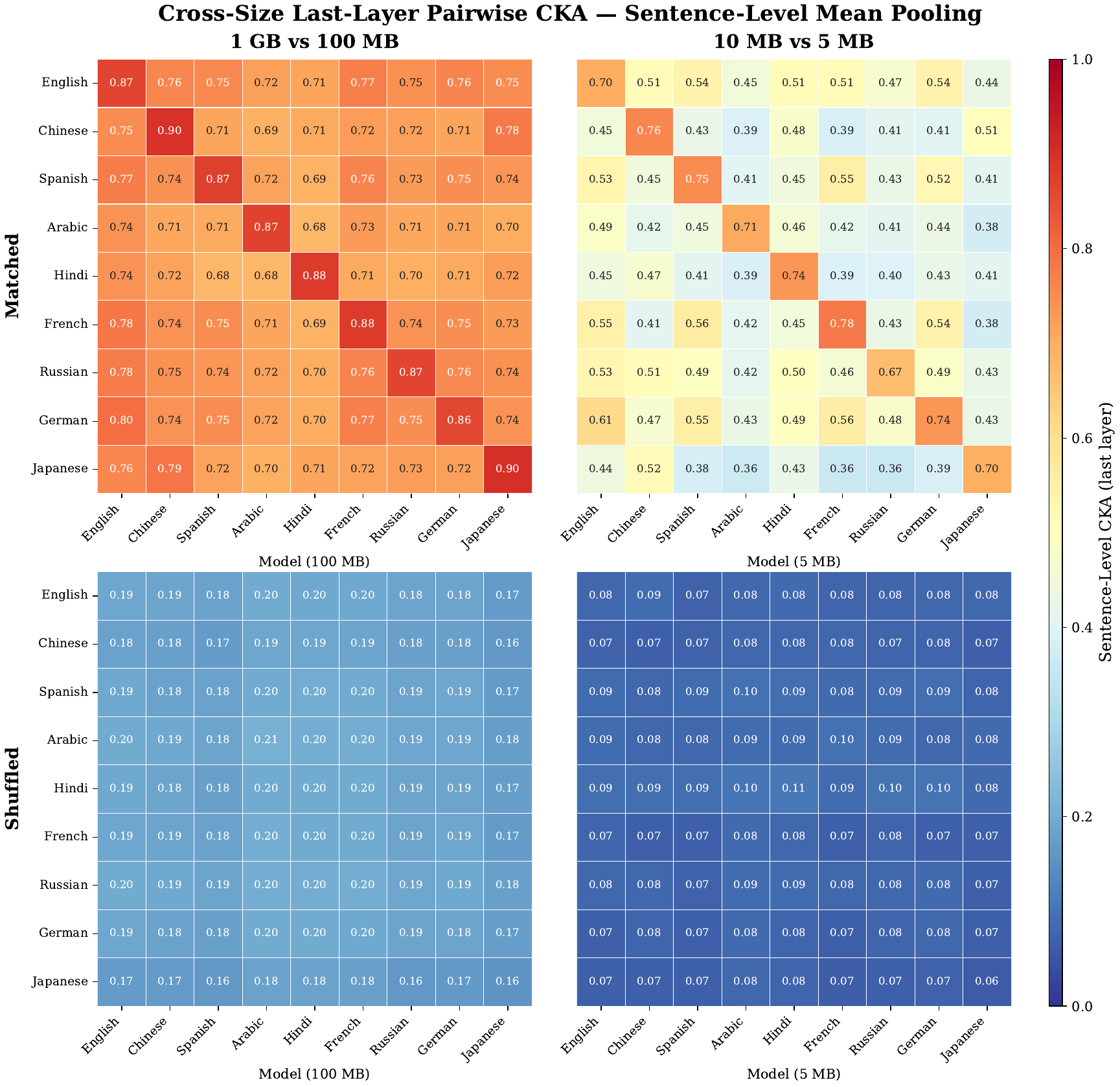}
    \caption{Sentence-Level Mean Pooling: cross-size CKA at the last layer. Left pair: 1000\,MB models (rows) vs.\ 100\,MB models (columns). Right pair: 10\,MB vs.\ 5\,MB. Within each pair, matched and shuffled are shown separately. Diagonal entries correspond to the \emph{same language} compared across sizes; off-diagonal entries correspond to \emph{different languages} at different scales.}
    \label{fig:cross_size_mean}
\end{figure}

\begin{figure}[h]
    \centering
    \includegraphics[width=\linewidth]{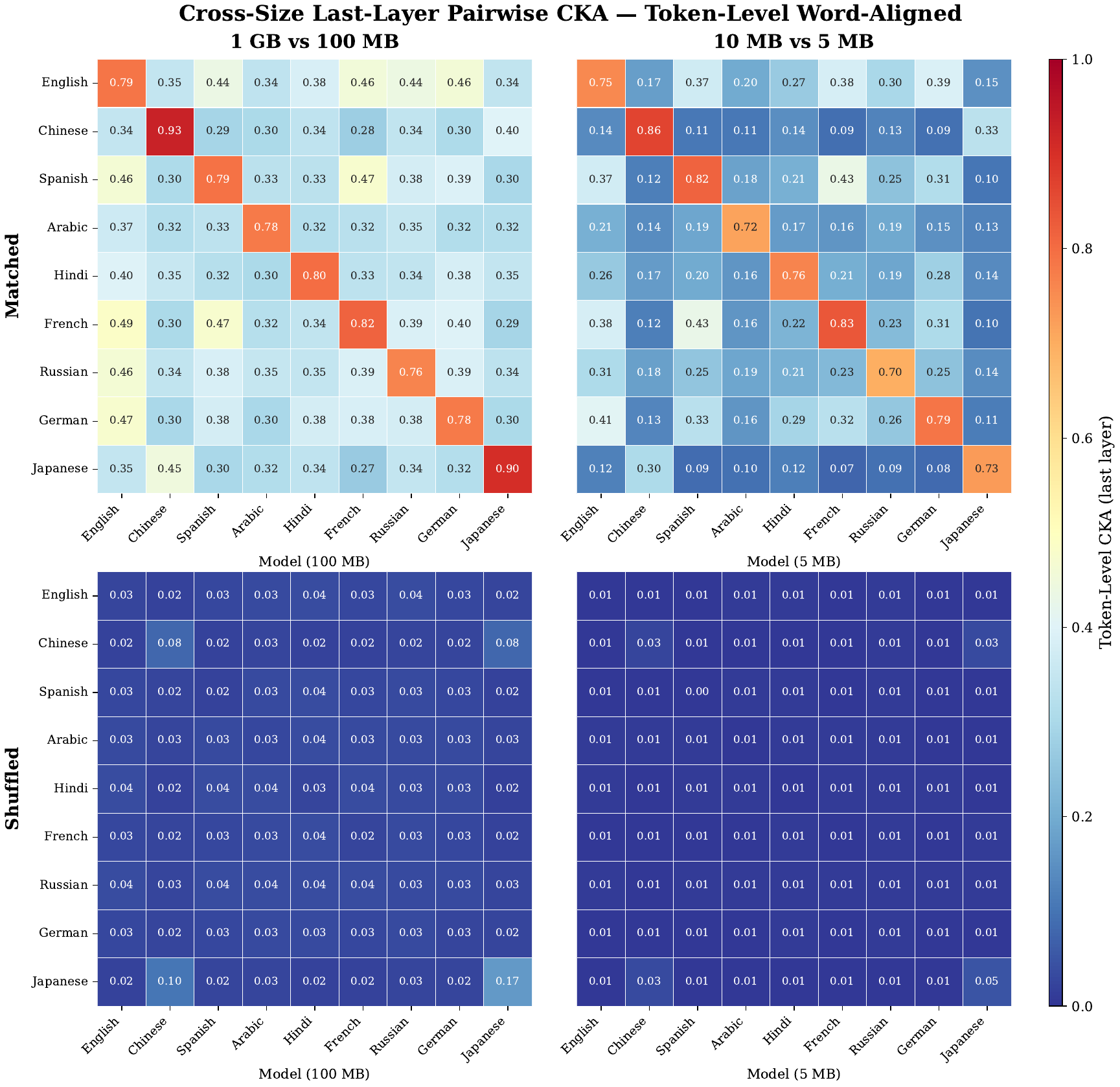}
    \caption{Token-Level Word-Aligned: cross-size CKA. Same layout as Figure~\ref{fig:cross_size_mean}.}
    \label{fig:cross_size_token}
\end{figure}

\begin{figure}[h]
    \centering
    \includegraphics[width=\linewidth]{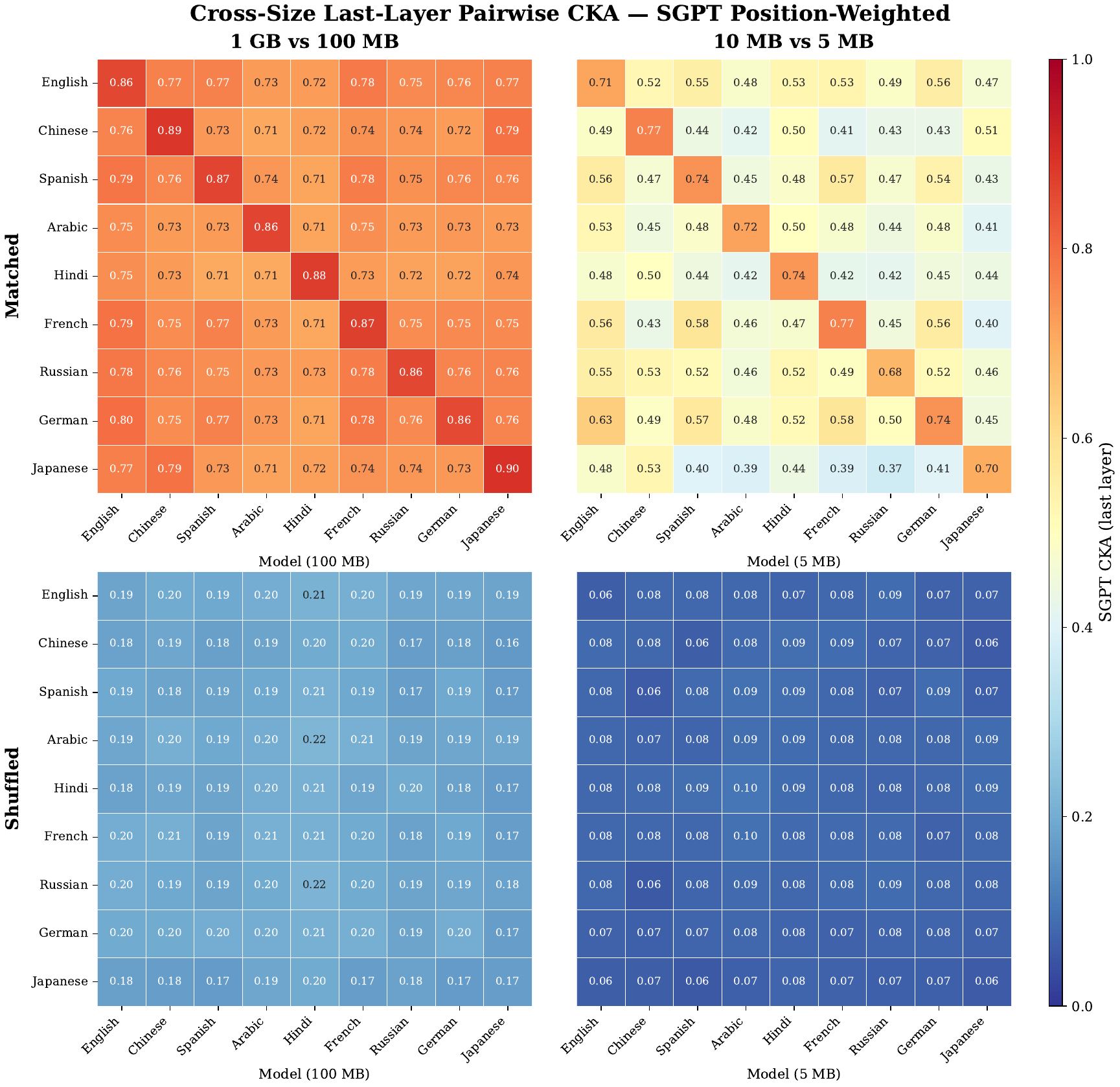}
    \caption{SGPT Position-Weighted Pooling: cross-size CKA. Same layout as Figure~\ref{fig:cross_size_mean}.}
    \label{fig:cross_size_sgpt}
\end{figure}

Figures~\ref{fig:cross_size_mean}--\ref{fig:cross_size_sgpt} show cross-size CKA matrices, where Model A and Model B are trained on \emph{different} amounts of data. We consider two cross-size pairings: 1000\,MB vs.\ 100\,MB, and 10\,MB vs.\ 5\,MB.

\paragraph{Findings.}
Two consistent patterns emerge across all three settings.

\textbf{(i) Same-language diagonal dominates.} When comparing, say, the English 1000\,MB model against the English 100\,MB model, the diagonal CKA is substantially higher than any off-diagonal cell in the same row.

\textbf{(ii) Matched signal survives across sizes.} In the 1000\,MB vs.\ 100\,MB comparison, off-diagonal matched CKA values remain clearly above the shuffled baseline, indicating that independently trained monolingual models at different scales still encode partially compatible concept representations. The 10\,MB vs.\ 5\,MB comparison shows lower absolute values (consistent with the within-size scaling result), but the matched-vs-shuffled gap persists.

\section{Cross-Architecture Comparison}
\label{sec:appendix_cross_arch}

This appendix supports Section~\ref{sec:cross_arch} with the full model specifications for both cross-architecture experiments, and a first-layer analysis that complements the last-layer matrix in the main text.

\subsection{Goldfish vs.\ Pythia}
\label{sec:appendix_cross_arch_pythia}

\begin{figure}[h]
\centering
\includegraphics[width=\linewidth]{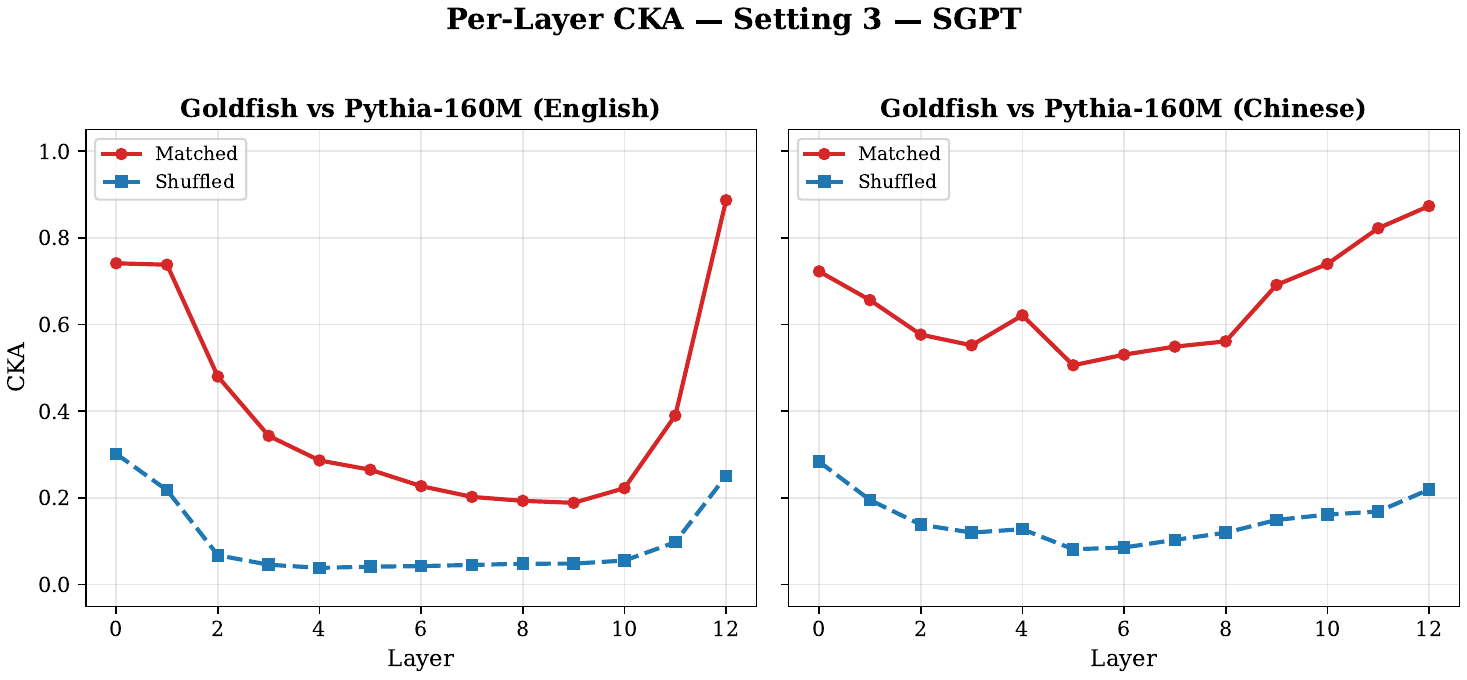}
\caption{\textbf{Goldfish vs.\ Pythia per-layer CKA (SGPT).} Left: English (Goldfish-en vs.\ Pythia-160M). Right: Chinese (Goldfish-zh vs.\ Zh-Pythia-160M). Red: matched; blue: shuffled.}
\label{fig:cross_arch_main}
\end{figure}

We compare Goldfish against Pythia-160M \citep{biderman2023pythia} (for English) and Zh-Pythia-160M \citep{liu2025systematicassessmentlanguagemodels} (for Chinese), which share Goldfish's hidden dimension and depth but differ in tokenizer, training data, and training pipeline. Figure~\ref{fig:cross_arch_main} reports per-layer SGPT CKA. Two observations stand out: matched CKA exceeds the shuffled baseline at \emph{every} layer for both English and Chinese, and the layerwise profile is U-shaped, with alignment high at embedding and final layers but depressed in the middle. This suggests that independently trained architectures converge on similar input- and output-facing semantic representations, but can diverge in their intermediate processing strategies. Table~\ref{tab:cross_arch_models} lists the four models used.

\begin{table}[h]
\centering
\small
\begin{tabular}{lll}
\toprule
\textbf{Model} & \textbf{Lang} & \textbf{Vocab} \\
\midrule
Goldfish 125M  & EN & 51\,200 \\
Pythia-160M    & EN & 50\,304 \\
\midrule
Goldfish 125M  & ZH & 51\,200 \\
Zh-Pythia-160M & ZH & 50\,304 \\
\bottomrule
\end{tabular}
\caption{Models used in the Goldfish vs.\ Pythia comparison of Section~\ref{sec:cross_arch}. All four models share hidden dimension 768, 12 layers, and 12 attention heads; they differ in vocabulary size (shown), tokenizer, training data, and training procedure.}
\label{tab:cross_arch_models}
\end{table}

\subsection{Diverse Independently-Developed Monolingual Models}
\label{sec:appendix_diverse_models}

Section~\ref{sec:cross_arch} (Table~\ref{tab:diverse_main}) reports last-layer SGPT CKA across five independently developed $\sim$1B monolingual models that share no architecture, tokenizer, training data, or research group. Here we provide the full model specifications and a first-layer comparison that complements the last-layer matrix in the main text.

\paragraph{Models.}
Table~\ref{tab:diverse_models} lists the five models. They span three architecture families (GPT-NeoX, LLaMA, Qwen2.5/Mistral), five languages, and five research groups, and differ in hidden dimension, depth, attention head count, and vocabulary.

\begin{table}[h]
\centering
\small
\setlength{\tabcolsep}{3pt}
\begin{tabular}{llcccc}
\toprule
\textbf{Model} & \textbf{Lang} & \textbf{Arch} & \textbf{Hidden} & \textbf{Layers} & \textbf{Heads} \\
\midrule
Pythia-1.4B     & EN & GPT-NeoX & 2048 & 24 & 16 \\
Zh-Pythia-1.4B  & ZH & GPT-NeoX & 2048 & 24 & 16 \\
Tucano-1B       & PT & LLaMA    & 2048 & 22 & 32 \\
Bielik-1.5B     & PL & Qwen2.5  & 1536 & 28 & 12 \\
Minerva-1B      & IT & Mistral  & 2048 & 16 & 16 \\
\bottomrule
\end{tabular}
\caption{Independently developed monolingual models used in the shared-nothing comparison of Section~\ref{sec:cross_arch}.}
\label{tab:diverse_models}
\end{table}

\paragraph{Setup.}
We compute SGPT-pooled linear CKA for all $\binom{5}{2}=10$ model pairs, with matched and shuffled controls. Because these models differ in depth, per-layer trajectories are not directly comparable; we therefore focus on the final pre-logit (last) layer, which represent the input and output boundaries of each model. Last-layer values are reported in Table~\ref{tab:diverse_main} of the main text.

\section{Predictors of Alignment: Additional Details}
\label{sec:appendix_explaining_alignment}

For each of the 36 Goldfish language pairs, we correlate last-layer SGPT matched CKA against five pairwise predictors: four URIEL typological distances \citep{littell-etal-2017-uriel} (syntactic, genetic, geographic, inventory) and the mean per-sentence negative log-likelihood (NLL) of the two models on FLORES, which proxies how well each model has learned its language. URIEL distances are taken directly from the URIEL typological database: syntactic distance over morphosyntactic feature vectors, genetic distance over the language family tree, geographic distance as great-circle distance between language centroids, and inventory distance over phonological inventories. For each predictor we compute Pearson and Spearman correlations against last-layer SGPT CKA (matched condition).

\begin{table}[h]
\centering
\small
\setlength{\tabcolsep}{4pt}
\begin{tabular}{lcccc}
\toprule
\textbf{Predictor} & $r$ & $p_r$ & $\rho$ & $p_\rho$ \\
\midrule
URIEL Syntactic  & $\mathbf{-0.603}$ & $<$\,.001 & $\mathbf{-0.635}$ & $<$\,.001 \\
URIEL Genetic    & $-$0.454 & .005 & $-$0.371 & .026 \\
Mean NLL         & $-$0.440 & .007 & $-$0.425 & .010 \\
URIEL Geographic & $-$0.387 & .020 & $-$0.440 & .007 \\
URIEL Inventory  & $-$0.209 & .222 & $-$0.123 & .477 \\
\bottomrule
\end{tabular}
\caption{\textbf{Predictors of last-layer matched CKA} across 36 Goldfish language pairs. $r$: Pearson; $\rho$: Spearman. $p_r$, $p_\rho$ are two-sided.}
\label{tab:alignment_predictors_main}
\end{table}

\begin{figure}[h]
\centering
\includegraphics[width=\linewidth]{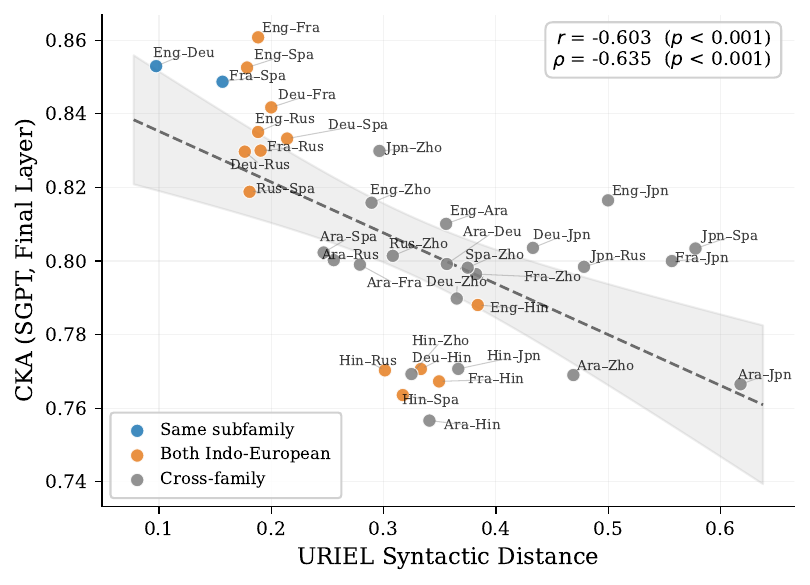}
\caption{\textbf{Last-layer matched CKA vs.\ URIEL syntactic distance} (SGPT) across 36 Goldfish pairs. Points are coloured by genealogical relationship: same subfamily (blue), both Indo-European (orange), cross-family (grey). Dashed line: linear fit with 95\% confidence band.}
\label{fig:cka_vs_syntactic_main}
\end{figure}

Syntactic distance is the strongest single predictor of alignment ($r=-0.60$, $\rho=-0.64$, $p<.001$), followed by genetic distance, mean NLL, and geographic distance; phonological inventory distance is not significantly correlated. Figure~\ref{fig:cka_vs_syntactic_main} visualises the syntactic relationship directly: same-subfamily pairs cluster in the low-distance, high-CKA quadrant, while cross-family pairs drift toward the opposite corner. The four URIEL distances are themselves intercorrelated, and we do not attempt to dissociate their contributions; the ranking is reported as a descriptive summary of where alignment is strongest. The negative coefficient on mean NLL is consistent with a model-fit effect: pairs whose models have learned their language less well also show weaker pairwise alignment.

\section{Training-Corpus Language Contamination Audit}
\label{sec:appendix_contamination}

Cross-lingual ability in nominally monolingual models can originate from foreign-language text in the training corpus \citep{blevins-zettlemoyer-2022-language}. We therefore audit the Goldfish training corpora used in this paper. For each language we sample 5{,}000 lines from the public training corpus and classify each line with GlotLID-v3 \citep{kargaran-etal-2023-glotlid}. Table~\ref{tab:contamination_audit} reports the fraction of lines labelled as the corpus language, the fraction confidently labelled as any other language (classifier confidence $\geq 0.9$), and the fraction labelled English.

\begin{table}[h]
\centering
\small
\setlength{\tabcolsep}{6pt}
\renewcommand{\arraystretch}{1.05}
\begin{tabular}{l rrr}
\toprule
\textbf{Corpus} & \textbf{Own (\%)} & \textbf{Other (\%)} & \textbf{English (\%)} \\
\midrule
eng & 99.7 & 0.00 & ---  \\
fra & 99.5 & 0.00 & 0.08 \\
rus & 99.6 & 0.10 & 0.00 \\
spa & 99.2 & 0.02 & 0.00 \\
deu & 97.4 & 0.10 & 0.02 \\
jpn & 98.3 & 0.06 & 0.00 \\
zho & 97.9 & 0.40 & 0.00 \\
hin & 96.3 & 0.82 & 0.10 \\
arb & 84.4$^{*}$ & 1.98 & 0.00 \\
\midrule
swh & 92.1 & 2.14 & 1.54 \\
tgl & 95.3 & 1.16 & 0.76 \\
amh & 93.5 & 4.64 & 0.58 \\
uzn & 97.3 & 1.00 & 0.18 \\
\bottomrule
\end{tabular}
\caption{GlotLID-v3 audit of the Goldfish training corpora, 5{,}000 sampled lines per language. Own: lines labelled as the corpus language. Other: lines labelled as any other language with confidence $\geq 0.9$. English: lines labelled English. $^{*}$The Arabic residual is almost entirely Standard Arabic assigned Arabic dialect labels (\texttt{ary}, \texttt{arz}, \texttt{ars}).}
\label{tab:contamination_audit}
\end{table}

For the nine languages of the main analysis, lines confidently identified as English are at most 0.1\% (French 0.08\%, Hindi 0.10\%, all others $\leq 0.02\%$). At the 1000\,MB tier this bounds incidental English exposure to roughly 1\,MB of text, below even the 5\,MB tier, the smallest at which we measure alignment (Section~\ref{sec:size_scaling}). The contamination mechanism of \citet{blevins-zettlemoyer-2022-language} requires meaningful high-resource leakage; at these rates it cannot supply the cross-lingual competence observed in Sections~\ref{sec:representational_alignment}--\ref{sec:crossmodel_patching}. The four low-resource corpora carry more foreign material (up to 2.14\% other-language and 1.54\% English for Swahili), consistent with noisier web sources; this is a caveat for Table~\ref{tab:cka_lowresource} and is noted in the Limitations. Line-level identification on web data is itself imperfect, particularly for closely related varieties \citep{suarez-etal-2026-commonlid}, which is visible in the Arabic dialect labelling above.

\section{Cross-Lingual Reconstruction: Method Comparison}
\label{app:projection_methods}

This appendix supports Section~\ref{sec:reconstruction} (\emph{Cross-Lingual Representation Reconstruction}) with extended results comparing the three projection families (Procrustes, Affine with ridge, MLP) on all 36 language pairs at the final layer.

\begin{figure}[h]
    \centering
    \includegraphics[width=\linewidth]{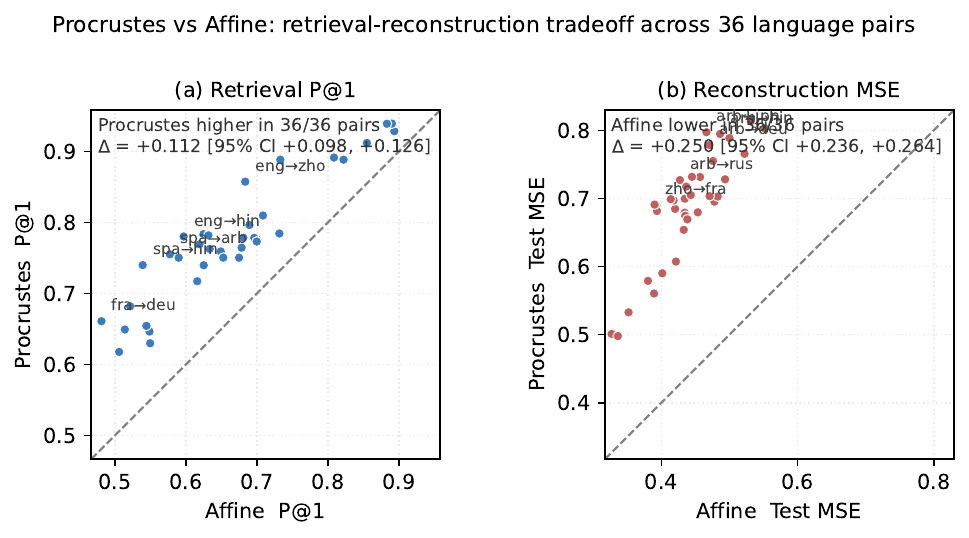}
    \caption{\textbf{Procrustes vs.\ Affine, all 36 pairs (layer 12).} Each point is one language pair. Left: retrieval P@1 against the train+test target pool; right: test MSE. Dashed line is $y=x$. Annotations report the paired mean difference $\Delta$ with bootstrap 95\% CI (10{,}000 resamples).}
    \label{fig:procrustes_vs_affine}
\end{figure}

\paragraph{Per-pair comparison.}
Figure~\ref{fig:procrustes_vs_affine} plots Procrustes against Affine on retrieval P@1 (left) and test MSE (right) for each of the 36 language pairs at layer 12. Procrustes wins on P@1 for every pair (paired mean $\Delta = +0.112$, 95\% CI $[+0.080, +0.143]$), while Affine wins on MSE for every pair. Together with Table~\ref{tab:projection_methods_summary}, this dissociation shows that the metric on which a method is ``best'' depends on the metric being optimised: Affine attains lower MSE because its hypothesis class is strictly larger, but the orthogonality constraint of Procrustes preserves angular structure that is better for cosine retrieval.

\paragraph{Aggregate summary.}
Table~\ref{tab:projection_methods_summary} (Appendix~\ref{sec:appendix_floats}) reports mean P@1, MSE, and CKA across the 36 pairs with bootstrap 95\% CIs, and the number of pairs on which each method achieves the best score for each metric.

\begin{figure*}[!htbp]
    \centering
    \includegraphics[width=\linewidth]{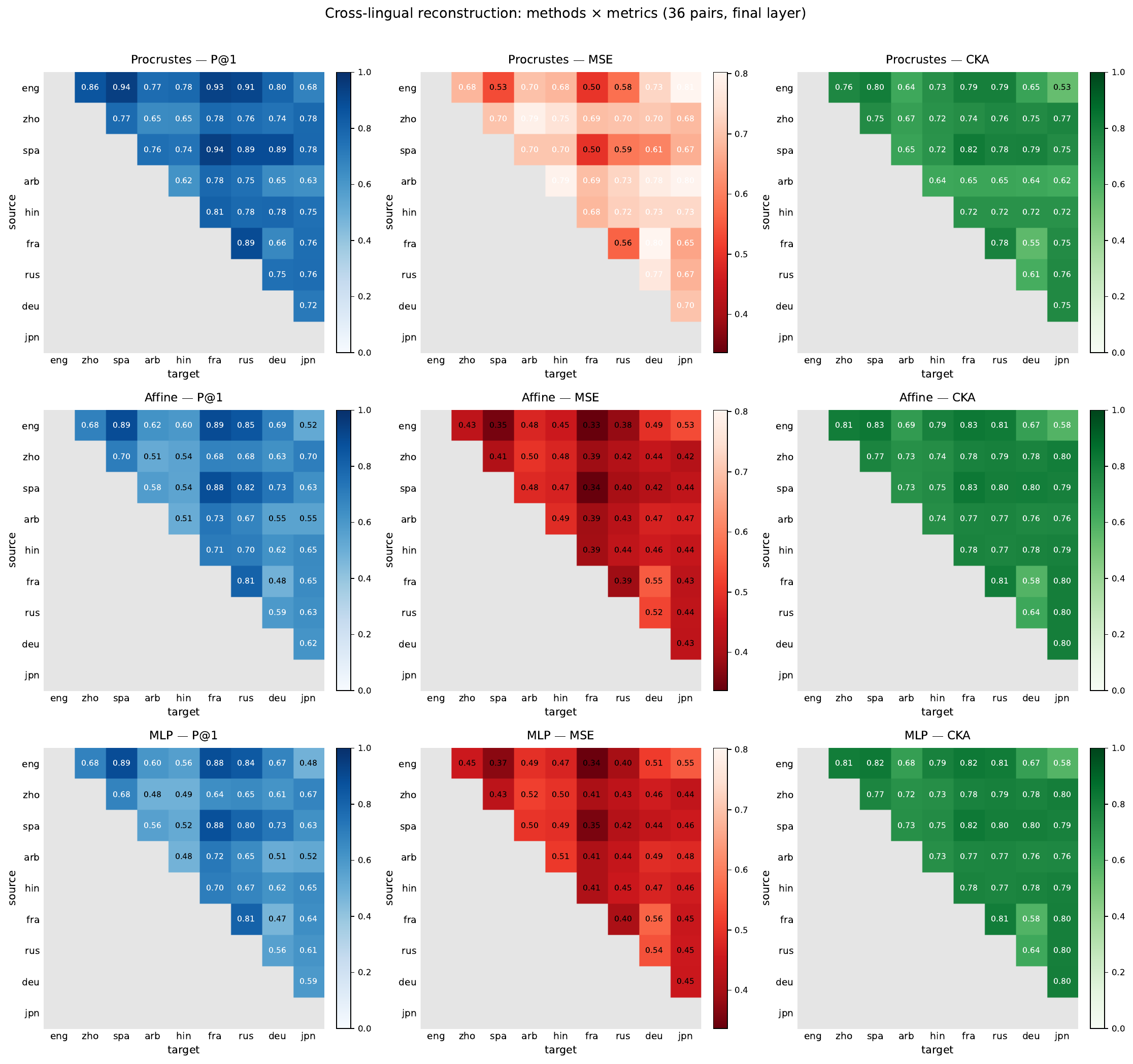}
    \caption{\textbf{Method$\times$metric heatmaps across layers and pairs.} Rows: methods (Procrustes, Affine, MLP). Columns: metrics (P@1, MSE, CKA). Each cell aggregates across 36 language pairs by layer.}
    \label{fig:methods_x_metrics}
\end{figure*}

\paragraph{Methods $\times$ metrics.}
Figure~\ref{fig:methods_x_metrics} (Appendix~\ref{sec:appendix_floats}) shows the full method$\times$metric grid: each panel is one (method, metric) combination across all 36 pairs and 13 layers. This confirms that the ranking observed at layer 12 is stable across most layers, and that MLP and Affine track each other closely, but Procrustes consistently dominates on retrieval.

\begin{figure*}[!htbp]
    \centering
    \includegraphics[width=\linewidth]{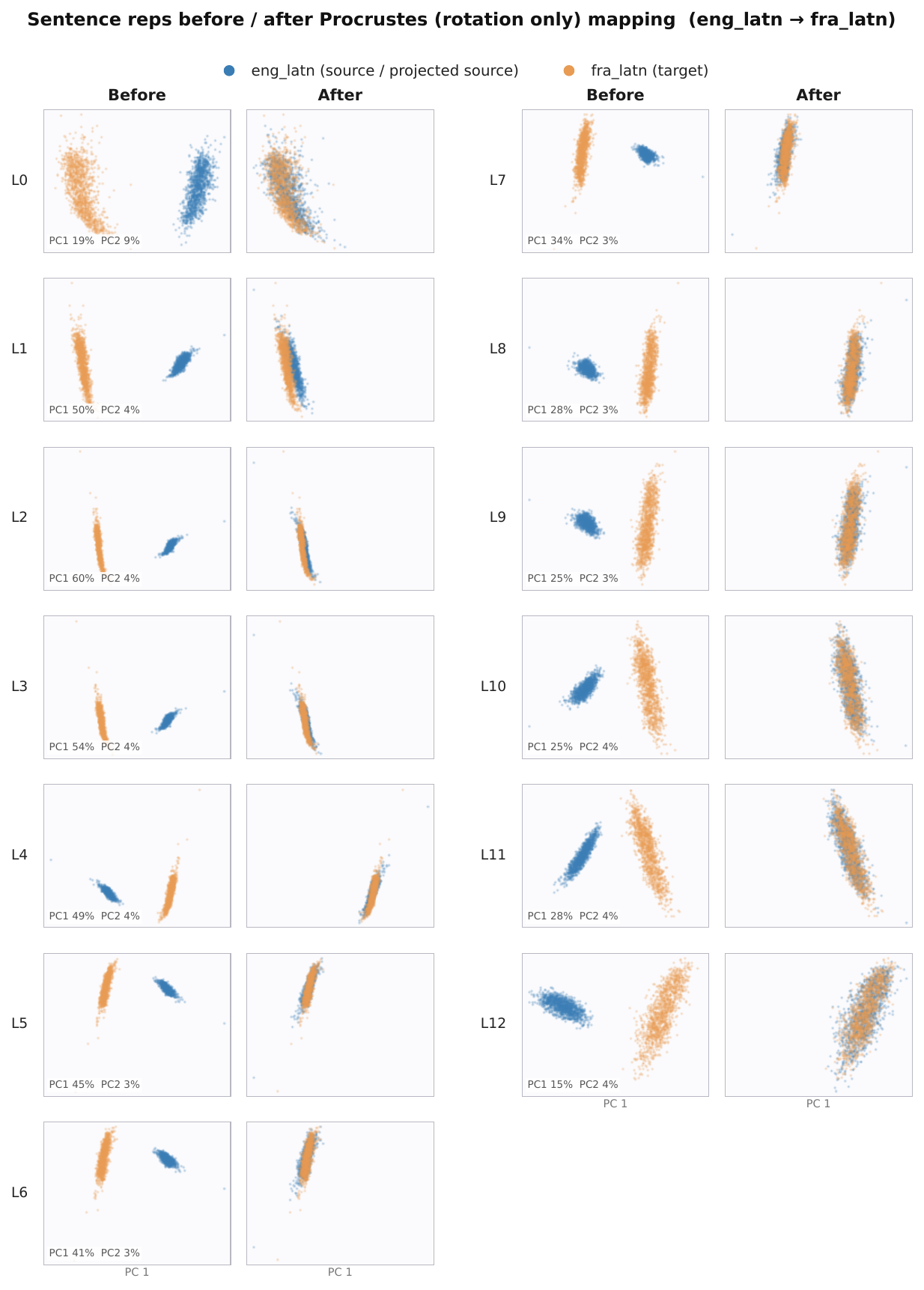}
    \caption{\textbf{Procrustes mapping, eng--fra} (peak retrieval layer; 2D PCA over the joint test set). English source ($\circ$), French target ($\bullet$), and Procrustes-mapped English ($\times$). Lines connect each English source to its matched French target.}
    \label{fig:pca_procrustes_engfra}
\end{figure*}

\paragraph{Visualising the rotation.}
Figure~\ref{fig:pca_procrustes_engfra} (Appendix~\ref{sec:appendix_floats}) provides a qualitative view of what Procrustes is doing. We project the English source representations and the matched French target representations into the same 2D PCA basis and overlay the Procrustes-mapped English points. This illustrates that a single rotation of the English space lands each sentence near its French counterpart.

\subsection{Layer-wise Analysis}
\label{sec:reconstruction_layers}

\begin{figure}[h]
\centering
\includegraphics[width=\linewidth]{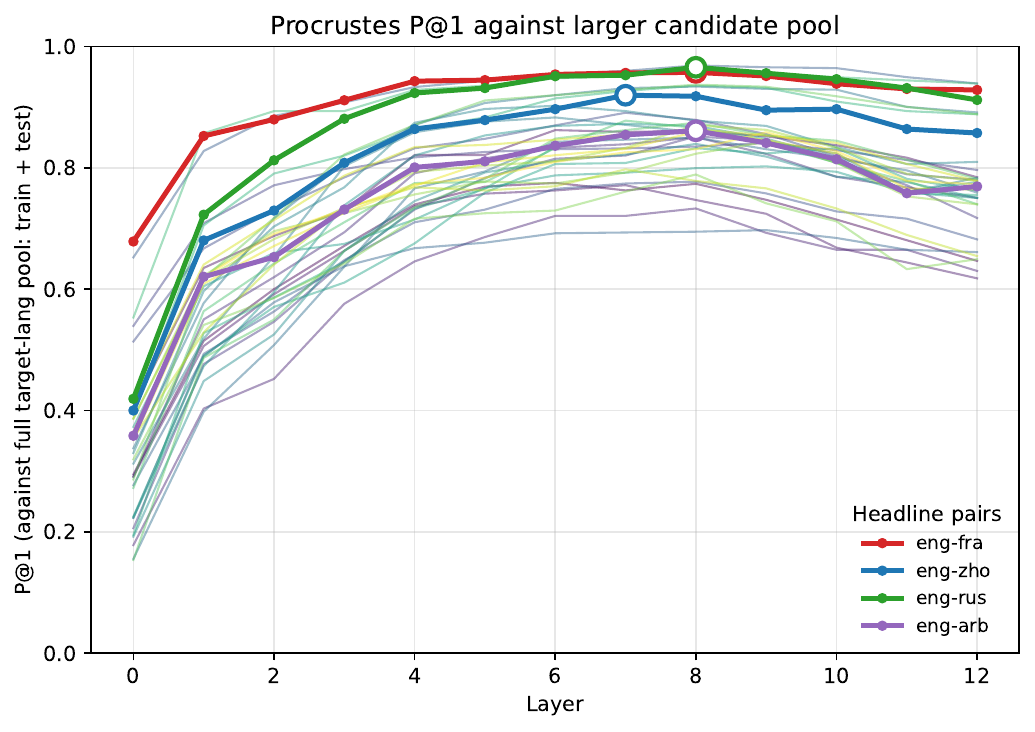}
\caption{\textbf{Procrustes P@1 by layer} (train${+}$test pool). Headline pairs (eng${\to}$\{fra, rus, zho, arb\}) bolded; remaining 32 pairs faint in background. Dots mark the per-pair best layer.}
\label{fig:procrustes_layers}
\end{figure}

We restrict to Procrustes and sweep layers $0$--$12$ for all 36 pairs. Figure~\ref{fig:procrustes_layers} reports retrieval P@1 by layer, with four representative pairs highlighted: eng${\to}$fra (typologically close), eng${\to}$rus (mid), eng${\to}$zho and eng${\to}$arb (distant). We see that retrieval is non-monotonic in depth and peaks at layers 6--8 across pairs (eng${\to}$fra: layer 8, P@1 ${=}0.981$; eng${\to}$rus: layer 6, $0.987$; eng${\to}$zho: layer 8, $0.977$; eng${\to}$arb: layer 8, $0.922$), with the location of the peak weakly tracking typological distance. The mid-layer peak is consistent with evidence that intermediate layers carry a more reliable multilingual signal than the final layer \citep{zhou2025beyondfinallayer}.

\subsection{Dimensional Analysis}
\label{sec:reconstruction_dim}

Procrustes leaves a residual. A natural follow-up question is what that residual looks like, because the answer determines whether a more flexible map could have closed the gap. Two possibilities could happen: Either the residual error is concentrated along a small number of directions, in which case the two spaces are nearly isometric except along those axes and a low-rank correction would suffice; or the error is spread across many directions, in which case no simple relaxation will help. We distinguish the two by taking an SVD of the residual and comparing its spectrum to that of the target.

\paragraph{Setup.}
Let $X,Y \in \mathbb{R}^{n\times d}$ be the per-layer source and target representations and $W^{\star}$ the Procrustes solution from Section~\ref{sec:reconstruction_setup}. The residual is $R = XW^{\star} - Y$. We take its SVD, $R = U_{R}\Sigma_{R}V_{R}^{\top}$, and the target's, $Y = U_{Y}\Sigma_{Y}V_{Y}^{\top}$, and compute two kinds of summary.

The first asks how spread-out a spectrum is. The entropy-based \emph{effective rank}
\[
\mathrm{eff\_rank}(M) = \exp\!\Big(-\!\sum_{i} \tilde{\sigma}_{i}\log\tilde{\sigma}_{i}\Big),\quad \tilde{\sigma}_{i} = \frac{\sigma_{i}^{2}}{\sum_{j}\sigma_{j}^{2}},
\]
counts the directions a matrix effectively occupies (a perfectly rank-1 matrix has eff\_rank $1$; a uniform spectrum across $d$ directions has eff\_rank $d$). We complement it with $k_\alpha$, the smallest $k$ whose top singular values carry an $\alpha$ fraction of the squared norm.

The second asks where the residual sits relative to the target. We project $R$ onto the top-$k$ right-singular vectors of $Y$, denoted $V_{Y}^{(k)}$, and report
\[
\mathrm{frac\_out}_{k} = 1 - \frac{\|R V_{Y}^{(k)}\|_{F}^{2}}{\|R\|_{F}^{2}},
\]
the fraction of residual energy lying outside the top-$k$ directions of $Y$. A small value means the residual lives in the same subspace the target uses; a large value means it lives off to the side, in directions the target itself barely populates.

\begin{table}[h]
\centering
\small
\setlength{\tabcolsep}{4pt}
\renewcommand{\arraystretch}{1.1}
\begin{tabular}{l cc cc cc}
\toprule
 & \multicolumn{2}{c}{\textbf{Eff.\ rank}} & \multicolumn{2}{c}{\textbf{Resid.\ $k_{\alpha}$}} & \multicolumn{2}{c}{\textbf{$\mathrm{frac\_out}_{k}$}} \\
\cmidrule(lr){2-3}\cmidrule(lr){4-5}\cmidrule(lr){6-7}
\textbf{Pair} & $Y$ & $R$ & $50\%$ & $90\%$ & $k{=}10$ & $k{=}50$ \\
\midrule
eng--fra & 67 & 246 & 90 & 357 & .93 & .73 \\
eng--rus & 45 & 202 & 74 & 273 & .93 & .73 \\
eng--zho & 43 & 180 & 69 & 265 & .91 & .69 \\
eng--arb & 53 & 147 & 71 & 306 & .88 & .68 \\
\bottomrule
\end{tabular}
\caption{\textbf{Spectra of $Y$ and the post-Procrustes residual $R$ at layer~8} ($d{=}768$). Columns: target and residual effective rank; smallest $k$ for which the top-$k$ singular values of $R$ capture $50\%$/$90\%$ of its squared norm; fraction of residual energy outside the top-$k$ right-singular subspace of $Y$.}
\label{tab:reconstruction_dim}
\end{table}

\paragraph{Findings.}
Table~\ref{tab:reconstruction_dim} reports these statistics for English paired with French, Russian, Chinese, and Arabic at the peak retrieval layer (layer~8). Three patterns hold across pairs and, with mild variation, across all 13 layers.

\textbf{(i) The target is low-rank; the residual is not.} The target representations use only $43$--$67$ directions out of $768$, and $90\%$ of their variance fits in roughly $160$--$200$ directions. The residual is far more diffuse: effective rank $147$--$246$, with half its energy already requiring $70$--$90$ singular values and $90\%$ requiring $265$--$360$. The error is therefore not stored in a small handful of bad axes that a targeted fix could correct; it is spread thinly across many directions.

\textbf{(ii) The residual lives outside the part of space the target uses.} Of the residual's total energy, $88$--$93\%$ lies outside the top-$10$ directions of $Y$, and $68$--$74\%$ outside the top-$50$. The Procrustes rotation lands cleanly in the subspace where the target stores its variance and leaves error in the directions the target barely populates. This is true even though the residual is large in absolute terms (Frobenius norm $0.79$--$0.93$ of the target's at layer~8): a residual can be large in raw magnitude while being functionally irrelevant if it sits where the model does not read.

\textbf{(iii) The pattern is stable across layers.} The same picture holds throughout the network. Residual effective rank exceeds target effective rank by a factor of at least two (typically three to six). $\mathrm{frac\_out}_{10}$ stays above $0.79$, and $\mathrm{frac\_out}_{50}$ above $0.57$, at every layer for every headline pair. No layer admits a low-dimensional collapse of the residual.

These observations reconcile a tension in the main results. Procrustes leaves a residual that is large in Frobenius norm yet small in functional consequence, because the residual is high-rank and largely orthogonal to the subspace where the target stores its variance. There is no concentrated low-dimensional structure for an Affine relaxation or a one-layer MLP to fit, which explains why neither improves retrieval over Procrustes. What the data support is approximate isometry along the dominant directions of the target, with diffuse mismatch in the remaining ones.

\section{Span Patching: Setup, Conditions, and Auxiliary Numbers}
\label{app:patching}

This appendix supports Section~\ref{sec:crossmodel_patching} with extended specification of the factual probes, the conditions, and the per-pair numbers underlying Figure~\ref{fig:patching_summary}.

\paragraph{Facts and prompts.}
The probe consists of 30 country$\to$capital facts, each available as a complete (prompt, answer) pair in six languages: English, French, German, Spanish, Japanese, and Chinese. Prompts share a fixed cloze form per language (e.g.\ \texttt{``The capital of $X$ is''},
 \texttt{``La capitale de $X$ est''},
 \texttt{``Die Hauptstadt von $X$ ist''},
 \texttt{``La capital de $X$ es''}, \texttt{``$X$\ja{の首都は}''}, \texttt{``$X$\zh{的首都是}''}). Answers are the target language's name for the capital, scored using each language's first subword id, computed in context so that BPE splits are handled correctly per tokenizer.

\paragraph{Span vs.\ single-layer patching.}
A preliminary experiment patched the residual at a single layer at the prompt's last-token position (eng$\to$fra at the position immediately before the LM head). Within-model $\Delta\log p(\text{donor})$ was $\sim$$+0.07$ nat, which doesn't seem to work well. This is becuase the intervention is too narrow: a single-layer last-token patch is immediately consumed by the LM head and downstream layers, with no opportunity to propagate. The span variant fixes this by (i) patching from $j$ to $L$, so each layer sees the patched residual at its input; and (ii) patching at the country (concept) position rather than the last token, so the patched information must move through attention to reach the answer logits.

\paragraph{Conditions.}
Four conditions per cross-fact cell are reported in the main results:

\begin{itemize}\itemsep0pt
    \item \texttt{within-lang}: target model, target-language donor prompt, identity projection. Within-model ceiling.
    \item \texttt{Procrustes}: source model on source-language donor prompt; per-layer orthogonal map $W^{(k)}$ from Section~\ref{sec:reconstruction}, fit on FLORES/OPUS/BouQUET sentence-end activations.
    \item \texttt{unprojected}: source-model residual at $c_S$, identity projection. Controls for basis mismatch.
    \item \texttt{shuffle}: Gaussian sample $\sim \mathcal{N}(\boldsymbol\mu_S^{(k)}, \boldsymbol\Sigma_S^{(k)})$ per layer with full-covariance Cholesky factorisation (eigendecomposition fallback for rank-deficient covariances), where $(\boldsymbol\mu_S^{(k)}, \boldsymbol\Sigma_S^{(k)})$ are the moments of the source-model activation pool. Distributional control.
\end{itemize}
A fifth condition, \texttt{Affine} (the per-layer linear$+$bias projection from Section~\ref{sec:reconstruction}, no orthogonality constraint), was evaluated and is discussed below; we exclude it from the main results for reasons later in this section.

\paragraph{Settings.}
Goldfish 1000\,MB models per language; 12 Transformer blocks; hidden dim 768; per-language vocabulary; max sequence length 512; seed 42. Span starts $j\in\{1,\dots,8\}$; span end fixed at $L=12$ (final block). Cross-fact pairs per (condition, $j$) cell: $30^{2}-30 = 870$ (excluding the diagonal where donor = target). The country position is identified per language by a verb marker preceding the noun phrase; the country's last subword is the token immediately before this marker. All cells run on a single GPU.

\paragraph{Multi-relation sweep.}
The relation sweep of Table~\ref{tab:patching_relations} uses the same span intervention between the English and German 1000\,MB models, with the patch start fixed at $j{=}2$ and the Section~\ref{sec:reconstruction} Procrustes maps applied without refitting. Cross-fact pairs combine each donor fact with a different target fact from the same relation, excluding the diagonal where donor equals target; $n$ varies with the facts available per relation. Several prompt-side and answer-side entities are multi-token under one or both tokenizers (e.g.\ \emph{Shakespeare} and \emph{Gutenberg} are three subwords in the German tokenizer; work titles such as \emph{Die Zauberfl\"ote} span several words); all conditions score $\Delta\log p$ at the answer's first subword, computed in context per tokenizer.

\begin{table*}[t]
\centering
\small
\setlength{\tabcolsep}{6pt}
\renewcommand{\arraystretch}{1.1}
\begin{tabular}{l r ccc}
\toprule
\textbf{Relation} & $n$ & \textbf{Within-lang} & \textbf{Procrustes} & \textbf{Shuffled} \\
\midrule
country$\to$capital    & 870 & 97.6 [96.3, 98.4] & 79.7 [76.9, 82.2] & 48.5 [45.2, 51.8] \\
city$\to$country       & 870 & 96.2 [94.7, 97.3] & 74.7 [71.7, 77.5] & 43.3 [40.1, 46.6] \\
country$\to$continent  & 398 & 87.9 [84.4, 90.8] & 53.3 [48.4, 58.1] & \phantom{0}5.3 [3.5, 7.9] \\
landmark$\to$city      & 450 & 76.7 [72.5, 80.3] & 60.7 [56.1, 65.1] & 38.9 [34.5, 43.5] \\
work$\to$author        & 132 & 54.5 [46.0, 62.8] & 47.0 [38.7, 55.4] & 17.4 [11.9, 24.8] \\
work$\to$composer      &  56 & 75.0 [62.3, 84.5] & 78.6 [66.2, 87.3] & 64.3 [51.2, 75.5] \\
invention$\to$inventor &  56 & 55.4 [42.4, 67.6] & 69.6 [56.7, 80.1] & 39.3 [27.6, 52.4] \\
\bottomrule
\end{tabular}
\caption{Directional success rate (\%) for English--German span patching across seven factual relations, with Wilson 95\% CIs in brackets. Same per-layer Procrustes maps as Figure~\ref{fig:patching_summary}, applied without refitting; patch span starts at $j{=}2$; answers scored at their first subword. $n$: cross-fact pairs per condition.}
\label{tab:patching_relations}
\end{table*}

\paragraph{Why Affine is excluded.}
We also evaluated the Section~\ref{sec:reconstruction} Affine projection on every eng$\to$X pair. Affine produces a magnitude artefact rather than directional transfer. Inspection of $\Delta_d{-}\Delta_t = \Delta\log p(\text{donor capital}) - \Delta\log p(\text{target capital})$, a stricter directional metric that requires the donor capital to rise \emph{more} than the target capital, is negative for Affine: it inflates the magnitude of the patched residual enough that the LM head boosts \emph{both} capitals, with the target capital often boosted more. The sign of $\Delta\log p(\text{donor capital})$ on Affine therefore reflects a uniform bias shift rather than functional transfer of the donor concept, so we report Procrustes as the cross-lingual projection of record and omit Affine from Figure~\ref{fig:patching_summary}.

\section{Supplementary Tables and Figures}
\label{sec:appendix_floats}

This appendix collects the wide tables and figures referenced from Appendices~\ref{sec:appendix_cka}, \ref{sec:appendix_model_size}, and \ref{app:projection_methods}.

\begin{table*}[!htbp]
\centering
\small
\caption{Last-layer pairwise CKA (\textit{matched\,/\,shuffled}) --- Sentence-Level Mean Pooling, FLORES dataset. All values computed on 1000\,MB Goldfish models.}
\label{tab:cka_mean_pool_flores}
\resizebox{\textwidth}{!}{%
\begin{tabular}{lccccccccc}
\toprule
 & \textbf{eng} & \textbf{zho} & \textbf{spa} & \textbf{arb} & \textbf{hin} & \textbf{fra} & \textbf{rus} & \textbf{deu} & \textbf{jpn} \\
\midrule
\textbf{eng} & --- & 0.80\,/\,0.22 & 0.84\,/\,0.23 & 0.80\,/\,0.24 & 0.78\,/\,0.23 & 0.85\,/\,0.24 & 0.82\,/\,0.24 & 0.85\,/\,0.24 & 0.81\,/\,0.21 \\
\textbf{zho} & 0.80\,/\,0.22 & --- & 0.78\,/\,0.21 & 0.75\,/\,0.23 & 0.76\,/\,0.22 & 0.78\,/\,0.22 & 0.79\,/\,0.23 & 0.77\,/\,0.23 & 0.82\,/\,0.20 \\
\textbf{spa} & 0.84\,/\,0.23 & 0.78\,/\,0.21 & --- & 0.78\,/\,0.24 & 0.74\,/\,0.23 & 0.83\,/\,0.24 & 0.80\,/\,0.24 & 0.82\,/\,0.24 & 0.78\,/\,0.21 \\
\textbf{arb} & 0.80\,/\,0.24 & 0.75\,/\,0.23 & 0.78\,/\,0.24 & --- & 0.73\,/\,0.23 & 0.78\,/\,0.25 & 0.78\,/\,0.25 & 0.78\,/\,0.25 & 0.75\,/\,0.22 \\
\textbf{hin} & 0.78\,/\,0.23 & 0.76\,/\,0.22 & 0.74\,/\,0.23 & 0.73\,/\,0.23 & --- & 0.75\,/\,0.24 & 0.75\,/\,0.24 & 0.76\,/\,0.24 & 0.77\,/\,0.21 \\
\textbf{fra} & 0.85\,/\,0.24 & 0.78\,/\,0.22 & 0.83\,/\,0.24 & 0.78\,/\,0.25 & 0.75\,/\,0.24 & --- & 0.82\,/\,0.24 & 0.83\,/\,0.24 & 0.78\,/\,0.21 \\
\textbf{rus} & 0.82\,/\,0.24 & 0.79\,/\,0.23 & 0.80\,/\,0.24 & 0.78\,/\,0.25 & 0.75\,/\,0.24 & 0.82\,/\,0.24 & --- & 0.82\,/\,0.25 & 0.79\,/\,0.22 \\
\textbf{deu} & 0.85\,/\,0.24 & 0.77\,/\,0.23 & 0.82\,/\,0.24 & 0.78\,/\,0.25 & 0.76\,/\,0.24 & 0.83\,/\,0.24 & 0.82\,/\,0.25 & --- & 0.79\,/\,0.21 \\
\textbf{jpn} & 0.81\,/\,0.21 & 0.82\,/\,0.20 & 0.78\,/\,0.21 & 0.75\,/\,0.22 & 0.77\,/\,0.21 & 0.78\,/\,0.21 & 0.79\,/\,0.22 & 0.79\,/\,0.21 & --- \\
\bottomrule
\end{tabular}}
\end{table*}

\begin{table*}[!htbp]
\centering
\small
\caption{Last-layer pairwise CKA (\textit{matched\,/\,shuffled}) --- Sentence-Level Mean Pooling, Tatoeba dataset. All values computed on 1000\,MB Goldfish models.}
\label{tab:cka_mean_pool_tatoeba}
\resizebox{\textwidth}{!}{%
\begin{tabular}{lccccccccc}
\toprule
 & \textbf{eng} & \textbf{zho} & \textbf{spa} & \textbf{arb} & \textbf{hin} & \textbf{fra} & \textbf{rus} & \textbf{deu} & \textbf{jpn} \\
\midrule
\textbf{eng} & --- & 0.45\,/\,0.17 & 0.78\,/\,0.21 & 0.64\,/\,0.22 & 0.65\,/\,0.26 & 0.75\,/\,0.23 & 0.70\,/\,0.22 & 0.81\,/\,0.23 & 0.59\,/\,0.25 \\
\textbf{zho} & 0.45\,/\,0.17 & --- & 0.60\,/\,0.23 & 0.51\,/\,0.31 & 0.50\,/\,0.20 & 0.57\,/\,0.21 & 0.53\,/\,0.21 & 0.63\,/\,0.26 & 0.43\,/\,0.16 \\
\textbf{spa} & 0.78\,/\,0.21 & 0.60\,/\,0.23 & --- & 0.68\,/\,0.24 & 0.85\,/\,0.61 & 0.72\,/\,0.23 & 0.70\,/\,0.23 & 0.74\,/\,0.24 & 0.62\,/\,0.27 \\
\textbf{arb} & 0.64\,/\,0.22 & 0.51\,/\,0.31 & 0.68\,/\,0.24 & --- & --- & 0.65\,/\,0.26 & 0.66\,/\,0.25 & 0.65\,/\,0.25 & 0.58\,/\,0.26 \\
\textbf{hin} & 0.65\,/\,0.26 & 0.50\,/\,0.20 & 0.85\,/\,0.61 & --- & --- & 0.83\,/\,0.46 & 0.86\,/\,0.66 & 0.84\,/\,0.58 & 0.74\,/\,0.33 \\
\textbf{fra} & 0.75\,/\,0.23 & 0.57\,/\,0.21 & 0.72\,/\,0.23 & 0.65\,/\,0.26 & 0.83\,/\,0.46 & --- & 0.67\,/\,0.21 & 0.75\,/\,0.25 & 0.53\,/\,0.25 \\
\textbf{rus} & 0.70\,/\,0.22 & 0.53\,/\,0.21 & 0.70\,/\,0.23 & 0.66\,/\,0.25 & 0.86\,/\,0.66 & 0.67\,/\,0.21 & --- & 0.73\,/\,0.26 & 0.65\,/\,0.27 \\
\textbf{deu} & 0.81\,/\,0.23 & 0.63\,/\,0.26 & 0.74\,/\,0.24 & 0.65\,/\,0.25 & 0.84\,/\,0.58 & 0.75\,/\,0.25 & 0.73\,/\,0.26 & --- & 0.61\,/\,0.26 \\
\textbf{jpn} & 0.59\,/\,0.25 & 0.43\,/\,0.16 & 0.62\,/\,0.27 & 0.58\,/\,0.26 & 0.74\,/\,0.33 & 0.53\,/\,0.25 & 0.65\,/\,0.27 & 0.61\,/\,0.26 & --- \\
\bottomrule
\end{tabular}}
\end{table*}

\begin{table*}[!htbp]
\centering
\small
\caption{Last-layer pairwise CKA (\textit{matched\,/\,shuffled}) --- Sentence-Level Mean Pooling, OPUS dataset. All values computed on 1000\,MB Goldfish models.}
\label{tab:cka_mean_pool_opus}
\resizebox{\textwidth}{!}{%
\begin{tabular}{lccccccccc}
\toprule
 & \textbf{eng} & \textbf{zho} & \textbf{spa} & \textbf{arb} & \textbf{hin} & \textbf{fra} & \textbf{rus} & \textbf{deu} & \textbf{jpn} \\
\midrule
\textbf{eng} & --- & 0.83\,/\,0.09 & 0.81\,/\,0.17 & 0.70\,/\,0.20 & 0.34\,/\,0.09 & 0.83\,/\,0.16 & 0.80\,/\,0.16 & 0.79\,/\,0.18 & 0.53\,/\,0.27 \\
\textbf{zho} & 0.83\,/\,0.09 & --- & --- & 0.74\,/\,0.13 & --- & 0.70\,/\,0.15 & 0.73\,/\,0.15 & 0.61\,/\,0.18 & --- \\
\textbf{spa} & 0.81\,/\,0.17 & --- & --- & --- & --- & --- & --- & --- & --- \\
\textbf{arb} & 0.70\,/\,0.20 & 0.74\,/\,0.13 & --- & --- & --- & 0.84\,/\,0.12 & 0.85\,/\,0.11 & 0.53\,/\,0.22 & --- \\
\textbf{hin} & 0.34\,/\,0.09 & --- & --- & --- & --- & --- & --- & --- & --- \\
\textbf{fra} & 0.83\,/\,0.16 & 0.70\,/\,0.15 & --- & 0.84\,/\,0.12 & --- & --- & 0.84\,/\,0.12 & 0.81\,/\,0.15 & --- \\
\textbf{rus} & 0.80\,/\,0.16 & 0.73\,/\,0.15 & --- & 0.85\,/\,0.11 & --- & 0.84\,/\,0.12 & --- & 0.62\,/\,0.24 & --- \\
\textbf{deu} & 0.79\,/\,0.18 & 0.61\,/\,0.18 & --- & 0.53\,/\,0.22 & --- & 0.81\,/\,0.15 & 0.62\,/\,0.24 & --- & --- \\
\textbf{jpn} & 0.53\,/\,0.27 & --- & --- & --- & --- & --- & --- & --- & --- \\
\bottomrule
\end{tabular}}
\end{table*}

\begin{table*}[!htbp]
\centering
\small
\caption{Last-layer pairwise CKA (\textit{matched\,/\,shuffled}) --- Sentence-Level Mean Pooling, BouQUET dataset. All values computed on 1000\,MB Goldfish models.}
\label{tab:cka_mean_pool_bouquet}
\resizebox{\textwidth}{!}{%
\begin{tabular}{lccccccccc}
\toprule
 & \textbf{eng} & \textbf{zho} & \textbf{spa} & \textbf{arb} & \textbf{hin} & \textbf{fra} & \textbf{rus} & \textbf{deu} & \textbf{jpn} \\
\midrule
\textbf{eng} & --- & 0.73\,/\,0.24 & 0.83\,/\,0.26 & 0.60\,/\,0.21 & 0.76\,/\,0.24 & 0.83\,/\,0.25 & 0.81\,/\,0.27 & 0.81\,/\,0.27 & 0.76\,/\,0.26 \\
\textbf{zho} & 0.73\,/\,0.24 & --- & 0.74\,/\,0.22 & 0.55\,/\,0.20 & 0.69\,/\,0.24 & 0.73\,/\,0.23 & 0.75\,/\,0.22 & 0.73\,/\,0.24 & 0.75\,/\,0.25 \\
\textbf{spa} & 0.83\,/\,0.26 & 0.74\,/\,0.22 & --- & 0.60\,/\,0.20 & 0.72\,/\,0.22 & 0.82\,/\,0.25 & 0.79\,/\,0.26 & 0.79\,/\,0.24 & 0.73\,/\,0.25 \\
\textbf{arb} & 0.60\,/\,0.21 & 0.55\,/\,0.20 & 0.60\,/\,0.20 & --- & 0.54\,/\,0.20 & 0.57\,/\,0.20 & 0.56\,/\,0.21 & 0.57\,/\,0.19 & 0.54\,/\,0.20 \\
\textbf{hin} & 0.76\,/\,0.24 & 0.69\,/\,0.24 & 0.72\,/\,0.22 & 0.54\,/\,0.20 & --- & 0.73\,/\,0.25 & 0.76\,/\,0.25 & 0.71\,/\,0.25 & 0.69\,/\,0.26 \\
\textbf{fra} & 0.83\,/\,0.25 & 0.73\,/\,0.23 & 0.82\,/\,0.25 & 0.57\,/\,0.20 & 0.73\,/\,0.25 & --- & 0.80\,/\,0.24 & 0.79\,/\,0.25 & 0.73\,/\,0.25 \\
\textbf{rus} & 0.81\,/\,0.27 & 0.75\,/\,0.22 & 0.79\,/\,0.26 & 0.56\,/\,0.21 & 0.76\,/\,0.25 & 0.80\,/\,0.24 & --- & 0.79\,/\,0.24 & 0.73\,/\,0.28 \\
\textbf{deu} & 0.81\,/\,0.27 & 0.73\,/\,0.24 & 0.79\,/\,0.24 & 0.57\,/\,0.19 & 0.71\,/\,0.25 & 0.79\,/\,0.25 & 0.79\,/\,0.24 & --- & 0.75\,/\,0.24 \\
\textbf{jpn} & 0.76\,/\,0.26 & 0.75\,/\,0.25 & 0.73\,/\,0.25 & 0.54\,/\,0.20 & 0.69\,/\,0.26 & 0.73\,/\,0.25 & 0.73\,/\,0.28 & 0.75\,/\,0.24 & --- \\
\bottomrule
\end{tabular}}
\end{table*}

\begin{table*}[!htbp]
\centering
\small
\caption{Last-layer pairwise CKA (\textit{matched\,/\,shuffled}) --- Token-Level Word-Aligned, FLORES dataset. All values computed on 1000\,MB Goldfish models.}
\label{tab:cka_token_aligned_flores}
\resizebox{\textwidth}{!}{%
\begin{tabular}{lccccccccc}
\toprule
 & \textbf{eng} & \textbf{zho} & \textbf{spa} & \textbf{arb} & \textbf{hin} & \textbf{fra} & \textbf{rus} & \textbf{deu} & \textbf{jpn} \\
\midrule
\textbf{eng} & --- & 0.37\,/\,0.03 & 0.51\,/\,0.04 & 0.40\,/\,0.04 & 0.43\,/\,0.04 & 0.54\,/\,0.04 & 0.51\,/\,0.04 & 0.52\,/\,0.03 & 0.37\,/\,0.03 \\
\textbf{zho} & 0.37\,/\,0.03 & --- & 0.32\,/\,0.03 & 0.33\,/\,0.03 & 0.37\,/\,0.03 & 0.32\,/\,0.03 & 0.36\,/\,0.03 & 0.31\,/\,0.03 & 0.43\,/\,0.10 \\
\textbf{spa} & 0.51\,/\,0.04 & 0.32\,/\,0.03 & --- & 0.38\,/\,0.04 & 0.36\,/\,0.04 & 0.53\,/\,0.03 & 0.43\,/\,0.04 & 0.43\,/\,0.04 & 0.32\,/\,0.03 \\
\textbf{arb} & 0.40\,/\,0.04 & 0.33\,/\,0.03 & 0.38\,/\,0.04 & --- & 0.35\,/\,0.04 & 0.38\,/\,0.04 & 0.40\,/\,0.04 & 0.35\,/\,0.04 & 0.34\,/\,0.03 \\
\textbf{hin} & 0.43\,/\,0.04 & 0.37\,/\,0.03 & 0.36\,/\,0.04 & 0.35\,/\,0.04 & --- & 0.38\,/\,0.04 & 0.38\,/\,0.04 & 0.42\,/\,0.04 & 0.37\,/\,0.02 \\
\textbf{fra} & 0.54\,/\,0.04 & 0.32\,/\,0.03 & 0.53\,/\,0.03 & 0.38\,/\,0.04 & 0.38\,/\,0.04 & --- & 0.44\,/\,0.04 & 0.44\,/\,0.03 & 0.31\,/\,0.03 \\
\textbf{rus} & 0.51\,/\,0.04 & 0.36\,/\,0.03 & 0.43\,/\,0.04 & 0.40\,/\,0.04 & 0.38\,/\,0.04 & 0.44\,/\,0.04 & --- & 0.43\,/\,0.04 & 0.36\,/\,0.03 \\
\textbf{deu} & 0.52\,/\,0.03 & 0.31\,/\,0.03 & 0.43\,/\,0.04 & 0.35\,/\,0.04 & 0.42\,/\,0.04 & 0.44\,/\,0.03 & 0.43\,/\,0.04 & --- & 0.32\,/\,0.03 \\
\textbf{jpn} & 0.37\,/\,0.03 & 0.43\,/\,0.10 & 0.32\,/\,0.03 & 0.34\,/\,0.03 & 0.37\,/\,0.02 & 0.31\,/\,0.03 & 0.36\,/\,0.03 & 0.32\,/\,0.03 & --- \\
\bottomrule
\end{tabular}}
\end{table*}

\begin{table*}[!htbp]
\centering
\small
\caption{Last-layer pairwise CKA (\textit{matched\,/\,shuffled}) --- Token-Level Word-Aligned, Tatoeba dataset. All values computed on 1000\,MB Goldfish models.}
\label{tab:cka_token_aligned_tatoeba}
\resizebox{\textwidth}{!}{%
\begin{tabular}{lccccccccc}
\toprule
 & \textbf{eng} & \textbf{zho} & \textbf{spa} & \textbf{arb} & \textbf{hin} & \textbf{fra} & \textbf{rus} & \textbf{deu} & \textbf{jpn} \\
\midrule
\textbf{eng} & --- & 0.17\,/\,0.04 & 0.51\,/\,0.07 & 0.38\,/\,0.07 & 0.39\,/\,0.09 & 0.51\,/\,0.06 & 0.51\,/\,0.06 & 0.50\,/\,0.05 & 0.30\,/\,0.08 \\
\textbf{zho} & 0.17\,/\,0.04 & --- & 0.27\,/\,0.07 & 0.33\,/\,0.12 & 0.23\,/\,0.05 & 0.22\,/\,0.04 & 0.24\,/\,0.06 & 0.25\,/\,0.06 & 0.44\,/\,0.19 \\
\textbf{spa} & 0.51\,/\,0.07 & 0.27\,/\,0.07 & --- & 0.47\,/\,0.11 & 0.61\,/\,0.28 & 0.54\,/\,0.06 & 0.48\,/\,0.08 & 0.43\,/\,0.06 & 0.30\,/\,0.09 \\
\textbf{arb} & 0.38\,/\,0.07 & 0.33\,/\,0.12 & 0.47\,/\,0.11 & --- & --- & 0.39\,/\,0.09 & 0.46\,/\,0.09 & 0.38\,/\,0.09 & 0.32\,/\,0.10 \\
\textbf{hin} & 0.39\,/\,0.09 & 0.23\,/\,0.05 & 0.61\,/\,0.28 & --- & --- & 0.56\,/\,0.19 & 0.58\,/\,0.27 & 0.55\,/\,0.20 & 0.42\,/\,0.10 \\
\textbf{fra} & 0.51\,/\,0.06 & 0.22\,/\,0.04 & 0.54\,/\,0.06 & 0.39\,/\,0.09 & 0.56\,/\,0.19 & --- & 0.52\,/\,0.05 & 0.46\,/\,0.05 & 0.26\,/\,0.07 \\
\textbf{rus} & 0.51\,/\,0.06 & 0.24\,/\,0.06 & 0.48\,/\,0.08 & 0.46\,/\,0.09 & 0.58\,/\,0.27 & 0.52\,/\,0.05 & --- & 0.48\,/\,0.07 & 0.33\,/\,0.08 \\
\textbf{deu} & 0.50\,/\,0.05 & 0.25\,/\,0.06 & 0.43\,/\,0.06 & 0.38\,/\,0.09 & 0.55\,/\,0.20 & 0.46\,/\,0.05 & 0.48\,/\,0.07 & --- & 0.29\,/\,0.07 \\
\textbf{jpn} & 0.30\,/\,0.08 & 0.44\,/\,0.19 & 0.30\,/\,0.09 & 0.32\,/\,0.10 & 0.42\,/\,0.10 & 0.26\,/\,0.07 & 0.33\,/\,0.08 & 0.29\,/\,0.07 & --- \\
\bottomrule
\end{tabular}}
\end{table*}

\begin{table*}[!htbp]
\centering
\small
\caption{Last-layer pairwise CKA (\textit{matched\,/\,shuffled}) --- Token-Level Word-Aligned, OPUS dataset. All values computed on 1000\,MB Goldfish models.}
\label{tab:cka_token_aligned_opus}
\resizebox{\textwidth}{!}{%
\begin{tabular}{lccccccccc}
\toprule
 & \textbf{eng} & \textbf{zho} & \textbf{spa} & \textbf{arb} & \textbf{hin} & \textbf{fra} & \textbf{rus} & \textbf{deu} & \textbf{jpn} \\
\midrule
\textbf{eng} & --- & 0.38\,/\,0.02 & 0.59\,/\,0.05 & 0.53\,/\,0.07 & 0.41\,/\,0.04 & 0.57\,/\,0.04 & 0.56\,/\,0.05 & 0.58\,/\,0.06 & 0.27\,/\,0.09 \\
\textbf{zho} & 0.38\,/\,0.02 & --- & --- & 0.35\,/\,0.02 & --- & 0.36\,/\,0.02 & 0.39\,/\,0.02 & 0.47\,/\,0.03 & --- \\
\textbf{spa} & 0.59\,/\,0.05 & --- & --- & --- & --- & --- & --- & --- & --- \\
\textbf{arb} & 0.53\,/\,0.07 & 0.35\,/\,0.02 & --- & --- & --- & 0.46\,/\,0.04 & 0.50\,/\,0.04 & 0.41\,/\,0.09 & --- \\
\textbf{hin} & 0.41\,/\,0.04 & --- & --- & --- & --- & --- & --- & --- & --- \\
\textbf{fra} & 0.57\,/\,0.04 & 0.36\,/\,0.02 & --- & 0.46\,/\,0.04 & --- & --- & 0.50\,/\,0.04 & 0.51\,/\,0.04 & --- \\
\textbf{rus} & 0.56\,/\,0.05 & 0.39\,/\,0.02 & --- & 0.50\,/\,0.04 & --- & 0.50\,/\,0.04 & --- & 0.45\,/\,0.07 & --- \\
\textbf{deu} & 0.58\,/\,0.06 & 0.47\,/\,0.03 & --- & 0.41\,/\,0.09 & --- & 0.51\,/\,0.04 & 0.45\,/\,0.07 & --- & --- \\
\textbf{jpn} & 0.27\,/\,0.09 & --- & --- & --- & --- & --- & --- & --- & --- \\
\bottomrule
\end{tabular}}
\end{table*}

\begin{table*}[!htbp]
\centering
\small
\caption{Last-layer pairwise CKA (\textit{matched\,/\,shuffled}) --- Token-Level Word-Aligned, BouQUET dataset. All values computed on 1000\,MB Goldfish models.}
\label{tab:cka_token_aligned_bouquet}
\resizebox{\textwidth}{!}{%
\begin{tabular}{lccccccccc}
\toprule
 & \textbf{eng} & \textbf{zho} & \textbf{spa} & \textbf{arb} & \textbf{hin} & \textbf{fra} & \textbf{rus} & \textbf{deu} & \textbf{jpn} \\
\midrule
\textbf{eng} & --- & 0.34\,/\,0.04 & 0.54\,/\,0.06 & 0.37\,/\,0.06 & 0.44\,/\,0.06 & 0.54\,/\,0.05 & 0.53\,/\,0.06 & 0.51\,/\,0.05 & 0.38\,/\,0.06 \\
\textbf{zho} & 0.34\,/\,0.04 & --- & 0.35\,/\,0.04 & 0.31\,/\,0.05 & 0.36\,/\,0.05 & 0.32\,/\,0.04 & 0.38\,/\,0.05 & 0.34\,/\,0.05 & 0.67\,/\,0.16 \\
\textbf{spa} & 0.54\,/\,0.06 & 0.35\,/\,0.04 & --- & 0.40\,/\,0.06 & 0.39\,/\,0.06 & 0.58\,/\,0.06 & 0.50\,/\,0.06 & 0.48\,/\,0.06 & 0.35\,/\,0.05 \\
\textbf{arb} & 0.37\,/\,0.06 & 0.31\,/\,0.05 & 0.40\,/\,0.06 & --- & 0.33\,/\,0.06 & 0.35\,/\,0.06 & 0.37\,/\,0.06 & 0.33\,/\,0.06 & 0.34\,/\,0.06 \\
\textbf{hin} & 0.44\,/\,0.06 & 0.36\,/\,0.05 & 0.39\,/\,0.06 & 0.33\,/\,0.06 & --- & 0.41\,/\,0.06 & 0.40\,/\,0.06 & 0.44\,/\,0.06 & 0.37\,/\,0.05 \\
\textbf{fra} & 0.54\,/\,0.05 & 0.32\,/\,0.04 & 0.58\,/\,0.06 & 0.35\,/\,0.06 & 0.41\,/\,0.06 & --- & 0.50\,/\,0.06 & 0.46\,/\,0.05 & 0.35\,/\,0.06 \\
\textbf{rus} & 0.53\,/\,0.06 & 0.38\,/\,0.05 & 0.50\,/\,0.06 & 0.37\,/\,0.06 & 0.40\,/\,0.06 & 0.50\,/\,0.06 & --- & 0.47\,/\,0.06 & 0.38\,/\,0.06 \\
\textbf{deu} & 0.51\,/\,0.05 & 0.34\,/\,0.05 & 0.48\,/\,0.06 & 0.33\,/\,0.06 & 0.44\,/\,0.06 & 0.46\,/\,0.05 & 0.47\,/\,0.06 & --- & 0.35\,/\,0.05 \\
\textbf{jpn} & 0.38\,/\,0.06 & 0.67\,/\,0.16 & 0.35\,/\,0.05 & 0.34\,/\,0.06 & 0.37\,/\,0.05 & 0.35\,/\,0.06 & 0.38\,/\,0.06 & 0.35\,/\,0.05 & --- \\
\bottomrule
\end{tabular}}
\end{table*}

\begin{table*}[!htbp]
\centering
\small
\caption{Last-layer pairwise CKA (\textit{matched\,/\,shuffled}) --- SGPT Position-Weighted Pooling, FLORES dataset. All values computed on 1000\,MB Goldfish models.}
\label{tab:cka_sgpt_flores}
\resizebox{\textwidth}{!}{%
\begin{tabular}{lccccccccc}
\toprule
 & \textbf{eng} & \textbf{zho} & \textbf{spa} & \textbf{arb} & \textbf{hin} & \textbf{fra} & \textbf{rus} & \textbf{deu} & \textbf{jpn} \\
\midrule
\textbf{eng} & --- & 0.82\,/\,0.23 & 0.85\,/\,0.24 & 0.81\,/\,0.24 & 0.79\,/\,0.24 & 0.86\,/\,0.24 & 0.84\,/\,0.25 & 0.85\,/\,0.24 & 0.82\,/\,0.23 \\
\textbf{zho} & 0.82\,/\,0.23 & --- & 0.80\,/\,0.22 & 0.77\,/\,0.23 & 0.77\,/\,0.23 & 0.80\,/\,0.24 & 0.80\,/\,0.23 & 0.79\,/\,0.24 & 0.83\,/\,0.21 \\
\textbf{spa} & 0.85\,/\,0.24 & 0.80\,/\,0.22 & --- & 0.80\,/\,0.23 & 0.76\,/\,0.23 & 0.85\,/\,0.23 & 0.82\,/\,0.23 & 0.83\,/\,0.24 & 0.80\,/\,0.22 \\
\textbf{arb} & 0.81\,/\,0.24 & 0.77\,/\,0.23 & 0.80\,/\,0.23 & --- & 0.76\,/\,0.25 & 0.80\,/\,0.25 & 0.80\,/\,0.25 & 0.80\,/\,0.24 & 0.77\,/\,0.24 \\
\textbf{hin} & 0.79\,/\,0.24 & 0.77\,/\,0.23 & 0.76\,/\,0.23 & 0.76\,/\,0.25 & --- & 0.77\,/\,0.24 & 0.77\,/\,0.25 & 0.77\,/\,0.24 & 0.77\,/\,0.22 \\
\textbf{fra} & 0.86\,/\,0.24 & 0.80\,/\,0.24 & 0.85\,/\,0.23 & 0.80\,/\,0.25 & 0.77\,/\,0.24 & --- & 0.83\,/\,0.24 & 0.84\,/\,0.24 & 0.80\,/\,0.22 \\
\textbf{rus} & 0.84\,/\,0.25 & 0.80\,/\,0.23 & 0.82\,/\,0.23 & 0.80\,/\,0.25 & 0.77\,/\,0.25 & 0.83\,/\,0.24 & --- & 0.83\,/\,0.25 & 0.80\,/\,0.23 \\
\textbf{deu} & 0.85\,/\,0.24 & 0.79\,/\,0.24 & 0.83\,/\,0.24 & 0.80\,/\,0.24 & 0.77\,/\,0.24 & 0.84\,/\,0.24 & 0.83\,/\,0.25 & --- & 0.80\,/\,0.22 \\
\textbf{jpn} & 0.82\,/\,0.23 & 0.83\,/\,0.21 & 0.80\,/\,0.22 & 0.77\,/\,0.24 & 0.77\,/\,0.22 & 0.80\,/\,0.22 & 0.80\,/\,0.23 & 0.80\,/\,0.22 & --- \\
\bottomrule
\end{tabular}}
\end{table*}

\begin{table*}[!htbp]
\centering
\small
\caption{Last-layer pairwise CKA (\textit{matched\,/\,shuffled}) --- SGPT Position-Weighted Pooling, Tatoeba dataset. All values computed on 1000\,MB Goldfish models.}
\label{tab:cka_sgpt_tatoeba}
\resizebox{\textwidth}{!}{%
\begin{tabular}{lccccccccc}
\toprule
 & \textbf{eng} & \textbf{zho} & \textbf{spa} & \textbf{arb} & \textbf{hin} & \textbf{fra} & \textbf{rus} & \textbf{deu} & \textbf{jpn} \\
\midrule
\textbf{eng} & --- & 0.43\,/\,0.17 & 0.81\,/\,0.21 & 0.65\,/\,0.22 & 0.66\,/\,0.25 & 0.76\,/\,0.24 & 0.72\,/\,0.23 & 0.81\,/\,0.23 & 0.63\,/\,0.24 \\
\textbf{zho} & 0.43\,/\,0.17 & --- & 0.62\,/\,0.22 & 0.52\,/\,0.33 & 0.48\,/\,0.19 & 0.59\,/\,0.22 & 0.52\,/\,0.21 & 0.64\,/\,0.24 & 0.40\,/\,0.15 \\
\textbf{spa} & 0.81\,/\,0.21 & 0.62\,/\,0.22 & --- & 0.69\,/\,0.24 & 0.87\,/\,0.65 & 0.76\,/\,0.23 & 0.74\,/\,0.24 & 0.76\,/\,0.24 & 0.65\,/\,0.26 \\
\textbf{arb} & 0.65\,/\,0.22 & 0.52\,/\,0.33 & 0.69\,/\,0.24 & --- & --- & 0.67\,/\,0.26 & 0.68\,/\,0.24 & 0.66\,/\,0.26 & 0.60\,/\,0.23 \\
\textbf{hin} & 0.66\,/\,0.25 & 0.48\,/\,0.19 & 0.87\,/\,0.65 & --- & --- & 0.83\,/\,0.51 & 0.85\,/\,0.63 & 0.85\,/\,0.57 & 0.75\,/\,0.32 \\
\textbf{fra} & 0.76\,/\,0.24 & 0.59\,/\,0.22 & 0.76\,/\,0.23 & 0.67\,/\,0.26 & 0.83\,/\,0.51 & --- & 0.70\,/\,0.23 & 0.76\,/\,0.24 & 0.60\,/\,0.26 \\
\textbf{rus} & 0.72\,/\,0.23 & 0.52\,/\,0.21 & 0.74\,/\,0.24 & 0.68\,/\,0.24 & 0.85\,/\,0.63 & 0.70\,/\,0.23 & --- & 0.74\,/\,0.26 & 0.68\,/\,0.26 \\
\textbf{deu} & 0.81\,/\,0.23 & 0.64\,/\,0.24 & 0.76\,/\,0.24 & 0.66\,/\,0.26 & 0.85\,/\,0.57 & 0.76\,/\,0.24 & 0.74\,/\,0.26 & --- & 0.64\,/\,0.26 \\
\textbf{jpn} & 0.63\,/\,0.24 & 0.40\,/\,0.15 & 0.65\,/\,0.26 & 0.60\,/\,0.23 & 0.75\,/\,0.32 & 0.60\,/\,0.26 & 0.68\,/\,0.26 & 0.64\,/\,0.26 & --- \\
\bottomrule
\end{tabular}}
\end{table*}

\begin{table*}[!htbp]
\centering
\small
\caption{Last-layer pairwise CKA (\textit{matched\,/\,shuffled}) --- SGPT Position-Weighted Pooling, OPUS dataset. All values computed on 1000\,MB Goldfish models.}
\label{tab:cka_sgpt_opus}
\resizebox{\textwidth}{!}{%
\begin{tabular}{lccccccccc}
\toprule
 & \textbf{eng} & \textbf{zho} & \textbf{spa} & \textbf{arb} & \textbf{hin} & \textbf{fra} & \textbf{rus} & \textbf{deu} & \textbf{jpn} \\
\midrule
\textbf{eng} & --- & 0.85\,/\,0.14 & 0.83\,/\,0.16 & 0.71\,/\,0.20 & 0.34\,/\,0.10 & 0.85\,/\,0.15 & 0.82\,/\,0.15 & 0.81\,/\,0.19 & 0.54\,/\,0.27 \\
\textbf{zho} & 0.85\,/\,0.14 & --- & --- & 0.74\,/\,0.14 & --- & 0.72\,/\,0.16 & 0.74\,/\,0.16 & 0.63\,/\,0.18 & --- \\
\textbf{spa} & 0.83\,/\,0.16 & --- & --- & --- & --- & --- & --- & --- & --- \\
\textbf{arb} & 0.71\,/\,0.20 & 0.74\,/\,0.14 & --- & --- & --- & 0.86\,/\,0.13 & 0.86\,/\,0.13 & 0.53\,/\,0.23 & --- \\
\textbf{hin} & 0.34\,/\,0.10 & --- & --- & --- & --- & --- & --- & --- & --- \\
\textbf{fra} & 0.85\,/\,0.15 & 0.72\,/\,0.16 & --- & 0.86\,/\,0.13 & --- & --- & 0.85\,/\,0.11 & 0.82\,/\,0.13 & --- \\
\textbf{rus} & 0.82\,/\,0.15 & 0.74\,/\,0.16 & --- & 0.86\,/\,0.13 & --- & 0.85\,/\,0.11 & --- & 0.63\,/\,0.23 & --- \\
\textbf{deu} & 0.81\,/\,0.19 & 0.63\,/\,0.18 & --- & 0.53\,/\,0.23 & --- & 0.82\,/\,0.13 & 0.63\,/\,0.23 & --- & --- \\
\textbf{jpn} & 0.54\,/\,0.27 & --- & --- & --- & --- & --- & --- & --- & --- \\
\bottomrule
\end{tabular}}
\end{table*}

\begin{table*}[!htbp]
\centering
\small
\caption{Last-layer pairwise CKA (\textit{matched\,/\,shuffled}) --- SGPT Position-Weighted Pooling, BouQUET dataset. All values computed on 1000\,MB Goldfish models.}
\label{tab:cka_sgpt_bouquet}
\resizebox{\textwidth}{!}{%
\begin{tabular}{lccccccccc}
\toprule
 & \textbf{eng} & \textbf{zho} & \textbf{spa} & \textbf{arb} & \textbf{hin} & \textbf{fra} & \textbf{rus} & \textbf{deu} & \textbf{jpn} \\
\midrule
\textbf{eng} & --- & 0.75\,/\,0.25 & 0.83\,/\,0.27 & 0.60\,/\,0.20 & 0.76\,/\,0.25 & 0.83\,/\,0.24 & 0.81\,/\,0.27 & 0.81\,/\,0.26 & 0.77\,/\,0.27 \\
\textbf{zho} & 0.75\,/\,0.25 & --- & 0.75\,/\,0.23 & 0.55\,/\,0.20 & 0.70\,/\,0.25 & 0.74\,/\,0.24 & 0.76\,/\,0.23 & 0.74\,/\,0.24 & 0.76\,/\,0.24 \\
\textbf{spa} & 0.83\,/\,0.27 & 0.75\,/\,0.23 & --- & 0.61\,/\,0.20 & 0.72\,/\,0.24 & 0.83\,/\,0.26 & 0.80\,/\,0.24 & 0.80\,/\,0.25 & 0.75\,/\,0.25 \\
\textbf{arb} & 0.60\,/\,0.20 & 0.55\,/\,0.20 & 0.61\,/\,0.20 & --- & 0.54\,/\,0.20 & 0.57\,/\,0.21 & 0.56\,/\,0.20 & 0.56\,/\,0.20 & 0.55\,/\,0.21 \\
\textbf{hin} & 0.76\,/\,0.25 & 0.70\,/\,0.25 & 0.72\,/\,0.24 & 0.54\,/\,0.20 & --- & 0.74\,/\,0.24 & 0.76\,/\,0.25 & 0.71\,/\,0.25 & 0.70\,/\,0.26 \\
\textbf{fra} & 0.83\,/\,0.24 & 0.74\,/\,0.24 & 0.83\,/\,0.26 & 0.57\,/\,0.21 & 0.74\,/\,0.24 & --- & 0.81\,/\,0.24 & 0.79\,/\,0.26 & 0.75\,/\,0.25 \\
\textbf{rus} & 0.81\,/\,0.27 & 0.76\,/\,0.23 & 0.80\,/\,0.24 & 0.56\,/\,0.20 & 0.76\,/\,0.25 & 0.81\,/\,0.24 & --- & 0.79\,/\,0.24 & 0.75\,/\,0.26 \\
\textbf{deu} & 0.81\,/\,0.26 & 0.74\,/\,0.24 & 0.80\,/\,0.25 & 0.56\,/\,0.20 & 0.71\,/\,0.25 & 0.79\,/\,0.26 & 0.79\,/\,0.24 & --- & 0.76\,/\,0.25 \\
\textbf{jpn} & 0.77\,/\,0.27 & 0.76\,/\,0.24 & 0.75\,/\,0.25 & 0.55\,/\,0.21 & 0.70\,/\,0.26 & 0.75\,/\,0.25 & 0.75\,/\,0.26 & 0.76\,/\,0.25 & --- \\
\bottomrule
\end{tabular}}
\end{table*}

\begin{table*}[!htbp]
\centering
\footnotesize
\setlength{\tabcolsep}{3pt}
\begin{tabular}{lllllll}
\toprule
Method & Mean P@1 & Mean MSE & Mean CKA & Best P@1 & Best MSE & Best CKA \\
\midrule
Procrustes & 0.775 [0.747, 0.803] & 0.689 [0.662, 0.715] & 0.713 [0.689, 0.735] & 36/36 & 0/36 & 0/36 \\
Affine & 0.663 [0.628, 0.699] & 0.439 [0.422, 0.455] & 0.762 [0.741, 0.781] & 0/36 & 36/36 & 32/36 \\
MLP & 0.643 [0.605, 0.682] & 0.456 [0.440, 0.473] & 0.760 [0.739, 0.778] & 0/36 & 0/36 & 4/36 \\
\bottomrule
\end{tabular}

\caption{\textbf{Method comparison across all 36 pairs at layer 12.} Cells show mean across 36 pairs with bootstrap 95\% CI (10{,}000 resamples). ``Best on metric'' counts the number of pairs on which each method achieves the top score.}
\label{tab:projection_methods_summary}
\end{table*}

\end{document}